\documentclass[final]{cvpr}
\usepackage{times}
\usepackage{epsfig}
\usepackage{graphicx}

\usepackage{amsmath}
\usepackage{amssymb}

\usepackage{multicol}
\usepackage{multirow}
\usepackage{array}
\usepackage{booktabs}
\usepackage{subcaption}
\usepackage{algorithm}
\usepackage{algpseudocode}
\usepackage{mathtools}
\usepackage[inline, shortlabels]{enumitem}
\usepackage[title]{appendix}
\setlist{itemjoin ={,\enspace},itemjoin* = { and\enspace}}
\newcommand{\R}{\mathbb{R}}

\def\mathanylap{\mathpalette\mathrlapinternal}
 \def\mathrlapinternal#1#2{%
 \rlap{$\mathsurround=26pt#1{#2}$}% $
}
\usepackage{subfiles}
\usepackage[pagebackref=true,breaklinks=true,colorlinks,bookmarks=false]{hyperref}

\newcolumntype{Z}[1]{>{\centering\arraybackslash}m{#1}}
\DeclareMathSymbol{\shortminus}{\mathbin}{AMSa}{"39}

\def\cvprPaperID{5623} % *** Enter the CVPR Paper ID here
\def\confYear{CVPR 2021}
\graphicspath{{images/}}

\begin{document}

%%%%%%%%% TITLE
\title{Fully Convolutional Scene Graph Generation}

\author{Hengyue Liu\textsuperscript{1}\thanks{Work done in part as an intern at Futurewei Technologies Inc.} \qquad\qquad
Ning Yan\textsuperscript{2} \qquad \qquad
Masood Mortazavi\textsuperscript{2} \qquad \qquad 
Bir Bhanu\textsuperscript{1}\\
\textsuperscript{1}University of California, Riverside \qquad 
\textsuperscript{2}Futurewei Technologies Inc.\\
\qquad \qquad {\tt\small \{hliu087, bir.bhanu\}@ucr.edu}, {\tt\small \{nyan, masood.mortazavi\}@futurewei.com}
}

\maketitle
% \thispagestyle{empty}
%%%%%%%%% ABSTRACT
\begin{abstract}
This paper presents a fully convolutional scene graph generation (FCSGG) model that detects objects and relations simultaneously. Most of the scene graph generation frameworks use a pre-trained two-stage object detector, like Faster R-CNN, and build scene graphs using bounding box features. Such pipeline usually has a large number of parameters and low inference speed. Unlike these approaches, FCSGG is a conceptually elegant and efficient bottom-up approach that encodes objects as bounding box center points, and relationships as 2D vector fields which are named as Relation Affinity Fields (RAFs). RAFs encode both semantic and spatial features, and explicitly represent the relationship between a pair of objects by the integral on a sub-region that points from subject to object. FCSGG only utilizes visual features and still generates strong results for scene graph generation. Comprehensive experiments on the Visual Genome dataset demonstrate the efficacy, efficiency, and generalizability of the proposed method. FCSGG achieves highly competitive results on recall and zero-shot recall with significantly reduced inference time.
\end{abstract}

%%%%%%%%% BODY TEXT
\section{Introduction}

Philosophers, linguists and artists have long wondered about the semantic content of what the mind perceives in images and speech~\cite{benjamin,chomsky_2006, poeticimage,rylemind}. Many have argued that images carry layers of meaning~\cite{berger,sontag}. Considered as an engineering problem, semantic content has been modeled either as latent representations~\cite{frome2013devise, h2020iclr,karpathy2015deep, mortazavi2020}, or explicitly as structured representations~\cite{lu2016visual,wu2019unified,xu2017scene}. For a computer vision system to explicitly represent and reason about the detailed semantics, Johnson~\cite{johnson2015image}~\etal adopt and formalize \emph{scene graphs} from computer graphics community. A scene graph serves as a powerful representation that enables many down-stream high-level reasoning tasks such as image captioning~\cite{yang2019auto,yao2018exploring}, image retrieval~\cite{johnson2018image,johnson2015image}, Visual Question answering~\cite{hudson2019gqa,teney2017graph} and image generation~\cite{johnson2018image,xu2018attngan}.

% where semantic proximity is estimated by the distance of vector representations in multi-dimensional ``latent'' spaces, ... Research on alignment or fusion of cross-modal features from vision and language has been widely investigated for both methods~\cite{h2020iclr, mortazavi2020, wu2019unified, xu2018attngan, johnson2018image, xu2017scene}.

% turn into the topic of SGG

\begin{figure}[t]
\centering
\includegraphics[width=1.0\linewidth, height=0.5\linewidth]{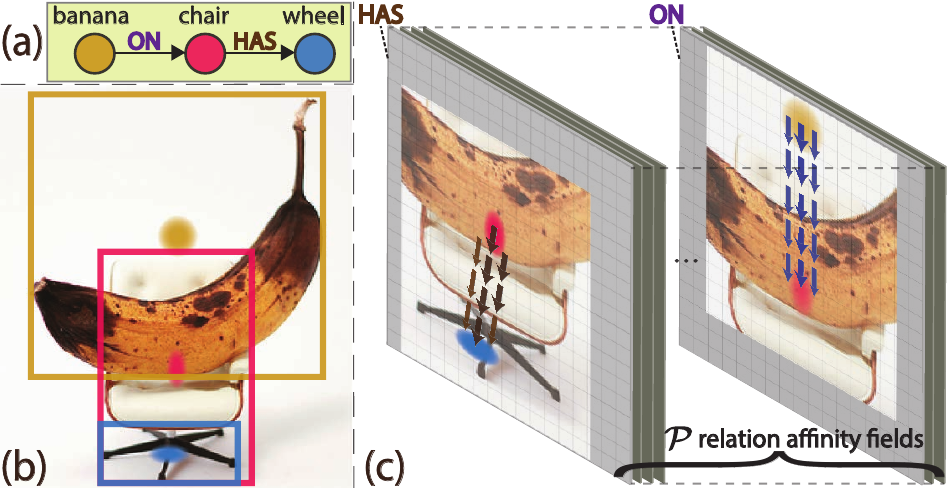}
\caption{An example of scene graph generation. (a) The ground-truth scene graph of an image. (b) The ground-truth bounding boxes and their centers. (c) Our proposed relationship representation called relation affinity fields. (The image is \texttt{2353896.jpg} from Visual Genome~\cite{krishna2017visual}.)}
\label{fig:teaser}
\end{figure}

A \emph{scene graph} is considered as an explicit structural representation for describing the semantics of a visual scene. The nodes in a scene graph represent the object classes and the edges represent the relationships between the objects. Figure~\ref{fig:teaser}(a) shows a simple example of a scene graph that represents the underlying semantics of an image. Each relationship between two objects is denoted as a triplet of \texttt{<}\texttt{subject}, \texttt{PREDICATE}, \texttt{object>} , i.e., \texttt{banana} $\xrightarrow{\texttt{ON}}$ \texttt{chair} and \texttt{chair} $\xrightarrow{\texttt{HAS}}$ \texttt{wheel} in Figure~\ref{fig:teaser}(a). Most of the SGG work~\cite{chen2019knowledge,tang2020unbiased,tang2019learning,xu2017scene,zellers2018neural} is build as a two-stage pipeline: object detection then scene graph generation. For the first stage of object detection, existing object detectors, \ie, Fast/Faster R-CNN~\cite{girshick2015fast,ren2015faster}, are used for object feature extraction from region proposals. For the second stage of scene graph generation, various approximation methods~\cite{tang2019learning,xu2017scene,zellers2018neural} for graph inference have been used. Some work~\cite{chen2019knowledge,yu2017visual,zareian2020bridging,zareian2020learning} have also investigated how to utilize external knowledge for improving the results. However, most previous work suffer from not only the long-tailed distribution of relationships~\cite{chen2019scene,dornadula2019visual,zellers2018neural}, but also the highly biased prediction conditioned on object labels~\cite{gkanatsiossaturation,hung2020contextual,knyazev2020graph,tang2020unbiased}. Consequently, frequent predicates will prevail over less frequent ones, and unseen relationships can not be identified. Moreover, the extensibility and inference speed of a SGG framework is crucial for accelerating down-stream tasks. Although few researchers have studied the efficiency and scalability in SGG~\cite{gkanatsios2019attention,li2018factorizable,yang2018graph}, the high computational complexity impedes the practicality towards real-world applications. 
A natural question that arises is: \emph{can we solve scene graph generation in a per-pixel prediction fashion?}
Recently, anchor-free object detectors~\cite{law2018cornernet,tian2019fcos,yang2019reppoints,zhou2019objects} have become popular due to their simplicity and low cost. These methods treat an object as a single or many, pre-defined or self-learned keypoints. Relating object detection to human pose estimation, if an object can be modeled as a point (human ``keypoint''), is it possible to represent a binary relationship as vectors (human ``limb'')?

%However, all previous results suffer from the long-tailed distribution of relationships where many relationships occur very few times in the dataset. Consequently, the generated scene graph edges are highly biased towards the more frequent relationships in the dataset. 
%%%% TODO: 
%%%% Masood's comment: The many sentences above have no references.
%%%% Many claims are made in thea bove few sentences without adequate references.
%%%% I know they are describing the traditiona/existing methods for SGG, it would 
%%%% still be useful to provide some references. 

In this paper, we propose a novel fully convolutional scene graph generation model, \ie, FCSGG, with state-of-the-art object detection results on Visual Genome dataset~\cite{krishna2017visual}, as well as compelling SGG results compared with visual-only methods. We present a bottom-up representation of objects and relationships by modeling objects as points and relationships as vectors. Each relationship is encoded as a segment in a 2D vector field called \emph{relation affinity field} (RAF). Figure~\ref{fig:teaser}(c) shows an illustration of RAFs for predicates \texttt{ON} and \texttt{HAS}. Both objects and relationships are predicted as dense feature maps without losing spatial information. For the first time, scene graphs can be generated from a single convolutional neural network (CNN) with significantly reduced model size and inference speed. Specifically, we make the following contributions:
\begin{itemize}
\setlength\itemsep{0em}
	\item We propose the first fully convolutional scene graph generation model that is more compact and computationally efficient compared to previous SGG models.
	\item We introduce a novel relationship representation called \emph{relation affinity fields} that generalizes well on unseen visual relationships. FCSGG achieves strong results on zero-shot recall.
	\item Our proposed model outperforms most of the visual-only SGG methods, and achieves competitive results compared to methods boosted by external knowledge.
	\item We conduct comprehensive experiments and benchmark our proposed method together with several previous work on model efficiency, and FCSGG achieves near real-time inference.
\end{itemize}

%The rest of paper is organized as follows. We discuss the related work in Section~\ref{sec:related}. We model the object detection task as a keypoint estimation problem in Section~\ref{sec:prelim}. We propose the concept of relation affinity fields for relationship representation and introduce our scene graph generation framework in Section~\ref{sec:rafs}. We discuss our experiments settings and results in Section~\ref{sec:experiments}. We conclude the paper in Section~\ref{sec:conclusions}. 

%------------------------------------------------------------------------

\section{Related Work}\label{sec:related}

%Follow CenterNet~\cite{zhou2019objects}, FCOS~\cite{tian2019fcos}, OpenPose~\cite{cao2019openpose}

% Hengyue's comments:
% basically we can divide the related work by categories, 
% methods with external knowledge vs. vision-only models
% some sub-topic could be
% feature refinement for SGG (RNN, LSTM, biLSTM, TreeLSTM...)
% utilize dataset statistics / long-tail property
% training / loss modifications
% design efficient SGG models
% we then stating our advantages and differentiate from each category

% we start another par on if relation should depending on the object:
% motifs dai2017detecting - conditional on object classes
% visual-only or object-agnostic work vip-cnn

% visual only yin2018zoom vtranse

% mentioned zero-shot lu2016visual plummer2017phrase yang2018shuffle hung2020contextual knyazev2020graph tang2020unbiased gkanatsiossaturation

% one-shot guo2020one

% few-shot dornadula2019visual

We categorize the related work of SGG into the following directions: refinement of contextual feature, adaptation of external knowledge, and others.

\emph{Contextual feature refinement}. Xu~\etal~\cite{xu2017scene} proposed an iterative message passing mechanism based on Gated Recurrent Units~\cite{cho2014properties}, where the hidden states are used for predictions. Followers~\cite{tang2019learning,zellers2018neural} studied better recurrent neural networks~\cite{hochreiter1997long,rumelhart1986learning,tai2015improved} for encoding object and edge context. Others trying to incorporate more spatial features into SGG. Li~\etal~\cite{li2017scene} proposed the MSDN that merges features across multiple semantic levels, and later achieved message passing constrained on visual phrase~\cite{li2017vip}. Dai~\etal\cite{dai2017detecting} proposed a spatial module by learning from bounding box masks. Woo~\etal~\cite{woo2018linknet} introduced the geometric embeddings by directly encoding the bounding box offsets between objects. Wang~\etal\cite{wang2019exploring} further studied the effects of relative positions between objects for extracting more discriminating features. Our method is fundamentally different from these methods as the relationships are grounded semantically and spatially directly into CNN features. Without any explicit iterative information exchange between nodes and edges, our model is able to predict objects and relationships in a single forward pass.

\emph{External knowledge adaptation}. Beyond visual features, linguistic knowledge can serve as additional features for SGG \cite{Gkanatsios_2019_ICCV,lu2016visual,plummer2017phrase,yu2017visual}. By adopting statistical correlations of objects, Chen~\etal~\cite{chen2019knowledge} utilized graph neural networks~\cite{scarselli2008graph} to infer relationships. Gu~\etal~\cite{gu2019scene} and Zareian~\etal\cite{zareian2020bridging, zareian2020learning} explored the usefulness of knowledge or commonsense graphs for SGG. Tang~\etal~\cite{tang2020unbiased} proposed an de-biasing method by causal interventions of predictions. Lin~\etal~\cite{lin2020gps} investigated the graph properties and mitigated the long-tailed distributions of relationships. Compared with these methods, our proposed model relies only on visual features but still yields a strong performance.

Very few researchers have investigated alternatives either for object feature or relationship feature representations. Newell~\cite{newell2017pixels} and Zhang~\etal~\cite{zhang2017visual} tried latent-space embeddings for relationship and achieved improvements. Different from most of the previous work, FCSGG reformulates and generalizes relationship representations from only visual-based features in near real-time, which is much faster than specifically designed SGG models for efficiency~\cite{li2018factorizable,yang2018graph}. 

%Due to the unbalanced nature of scene graph datasets, many researchers work on semi-supervised learning~\cite{chen2019scene} and few-shot learning~\cite{dornadula2019visual,guo2020one} of SGG problem.

%------------------------------------------------------------------------

\section{Object Detection as Keypoint Estimation} \label{sec:prelim}

%----- Architecture diagram 
\begin{figure*}[t]
\centering
\includegraphics[width=\textwidth,height=0.28\textwidth]{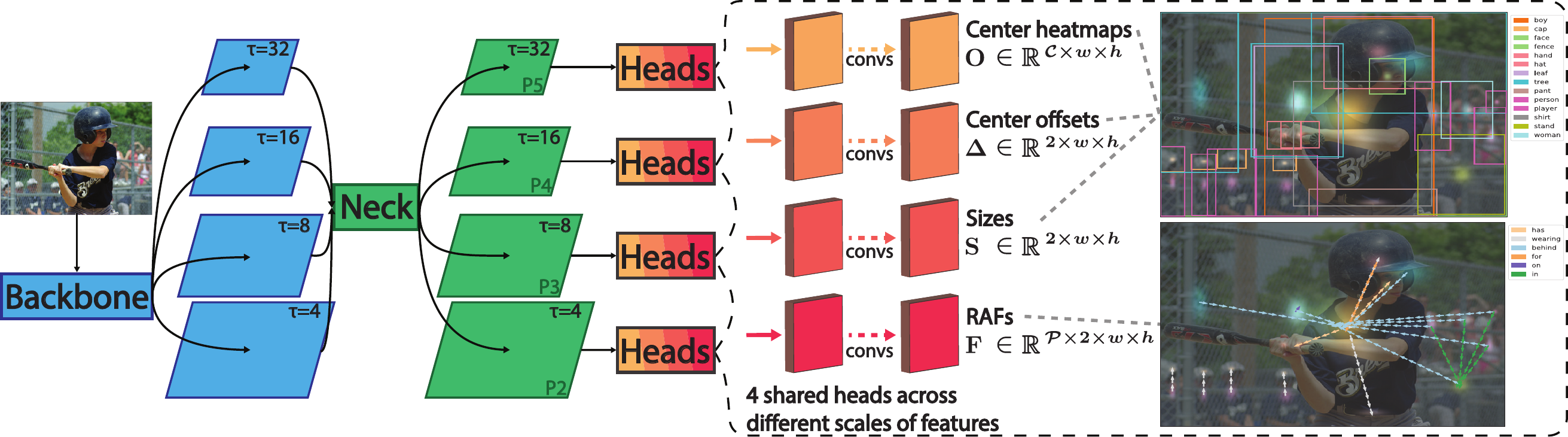}
\caption{One example of our proposed fully convolutional scene graph generation architecture using four scales of features for prediction. We refer to the ``backbone'' as the feature extraction CNN like ResNet~\cite{he2016deep}, and the ``neck'' as the network for generating multi-scale features like FPN~\cite{lin2017feature}, and the head as several convolutional layers (\textsf{convs} in figure). Shown in the right part, there are four output features per scale: $\mathbf{O}$, $\mathbf{\Delta}$, $\mathbf{S}$ for object detection and $\mathbf{F}$ for relationship detection. For single-scale prediction, the backbone features of $\tau=4$ will be directly fed into the heads.} %Please zoom in for better visualizations.
\label{fig:architecture}
\end{figure*}

In this section, we provide the preliminaries of modeling object detection as keypoint estimation in a single-scale dense feature map prediction fashion. 

Our model is built upon a one-stage anchor-free detector, namely CenterNet~\cite{zhou2019objects}. Different from commonly used anchor-based R-CNN approaches for generating object proposals and features, it predicts three dense features that represent centers of object bounding boxes, center offsets, and object sizes. More specifically, an input image $\mathrm{I} \in \R^{3 \times W \times H}$ will go through a backbone CNN generating feature maps of size $w \times h = \lfloor \frac{W}{\tau} \rfloor \times \lfloor \frac{H}{\tau} \rfloor $ where $\tau$ is the total stride until the last layer, and we set $\tau = 4$ unless specified otherwise. Then these features will be fed into three prediction heads, each of which consists of several convolutional layers. The three heads are for predicting object center heatmaps $\mathbf{O} \in \R^{\mathcal{C} \times w \times h}$ where $\mathcal{C}$ is the number of object classes in a dataset, object center offsets $\mathbf{\Delta} \in \R^{2 \times w \times h}$ for recovering from downsampled coordinates, and object sizes $\mathbf{S} \in \R^{2 \times w \times h}$, respectively (shown in the dashed block of Figure~\ref{fig:architecture}). We define the ground-truth (GT) objects in an image as $\mathbf{B} = \{ b^{i} \}$ where $ b^{i} = (x_0^{i}, y_0^{i}, x_1^{i}, y_1^{i}, c^{i})$ is the object $i$ of class $c^{i}$, $(x_0^{i}, y_0^{i})$ and $(x_1^{i}, y_1^{i})$ denote the coordinates of the left-top and right-bottom corners of its bounding box. The center of the bounding box is defined as $\mathbf{o}^{i} = (\mathbf{o}_x^{i}, \mathbf{o}_y^{i}) = ( (x_0^{i}+x_1^{i}) / 2, (y_0^{i}+y_1^{i}) / 2 )$, and the size of the object is defined as $ \mathbf{s}^{i} = ( x_1^{i} - x_0^{i}, y_1^{i} - y_0^{i} ) $. To obtain the ground-truth center heatmaps at feature level, we divide coordinates by the stride $\tau$ and add Gaussian-smoothed samples following Law~\cite{law2018cornernet} and Zhou~\etal~\cite{zhou2019objects}. Formally, the object center $\mathbf{o}^{i}$ will be modulated by a bivariate Gaussian distribution along x-axis and y-axis on $\mathbf{O}_{c^i}$. The value around $\mathbf{o}^{i}$ is computed as
\begin{equation}
%\resizebox{\linewidth}{!}{$
	\mathbf{O}_{c^i,x,y} = 
	\exp 
	\big( 
		\shortminus \displaystyle\frac{\|x\shortminus\lfloor \mathbf{o}^{i}_{x} / \tau \rfloor \|_{2}^2}{2\sigma_{x}^2} 
		\shortminus \frac{\|y\shortminus \lfloor \mathbf{o}^{i}_{y} / \tau \rfloor \|_{2}^2}{2\sigma_{y}^2}
	\big),
%$}
\label{eq:heatmap}
\end{equation}
where $\sigma_{x}$ and $\sigma_{y}$ controls the spread of the distribution.
When multiple objects of the same class $c$ contribute to $\mathbf{O}_{c,x,y}$, the maximum is taken as the ground truth. 
The center heatmaps are then supervised by Gaussian focal loss~\cite{law2018cornernet,lin2017focal,zhou2019objects}. More details are provided in the supplementary file.

In addition to the supervision of center heatmaps, the center offset regression $\mathcal{L}_{\mathbf{\Delta}}$ and object size regression $\mathcal{L}_{\mathbf{S}}$ are used to recover object detections. For mitigating discretization error due to downsampling, the offset target is $ \boldsymbol{\delta}^{i} = \mathbf{o}^{i} / \tau - \lfloor \mathbf{o}^{i} / \tau \rfloor $, and regressed via L1 loss as $\mathcal{L}_{\mathbf{\Delta}_{x,y}} \text{=\;} ||\hat{\mathbf{\Delta}}_{x,y} - \boldsymbol{\delta}^{i}||_{1}$ at center locations. For object size regression $\mathcal{L}_{\mathbf{S}}$, the target is feature-level object size $ \mathbf{s}^{i} / \tau $, and the actual size can be recovered by multiplying the output stride. We also use L1 loss as $\mathcal{L}_{\mathbf{S}_{x,y}} \text{=\;} ||\mathbf{\hat{S}}_{x,y} - \mathbf{s}^{i}||_{1}$ at center locations. Both object size and offset regressors are class-agnostic such that there will be only one valid regression target at a particular location where $\mathbf{O}_{c,x,y}=1$. If two object centers collide onto the same location, we choose the smaller object for regression. The overall object detection objective is
\begin{equation}
	\mathcal{L}_{det} = \frac{1}{N} \sum\limits_{c,x,y}
	\left(
		\mathcal{L}_{\mathbf{O}_{c,x,y}}  +\lambda_{\mathbf{\Delta}}\mathcal{L}_{\mathbf{\Delta}_{x,y}}
		+ \lambda_{\mathbf{s}}\mathcal{L}_{\mathbf{S}_{x,y}}
	\right),
\label{eq:overall}
\end{equation}
where $N$ is the total number of objects in the image, $\lambda_{\mathbf{\Delta}}$ and $\lambda_\mathbf{s}$ are hyper-parameters for weight balancing. We empirically set $\lambda_{\mathbf{\Delta}} = 1$ and $\lambda_{\mathbf{s}} = 0.1$ for all experiments. Until here, object centers, offsets, and sizes are all represented in single-scale feature maps. We will discuss a multi-scale prediction approach reducing regression ambiguity effectively in section~\ref{sec:ms}.

%------------------------------------------------------------------------

\section{Relation Affinity Fields} \label{sec:rafs}

Newell and Deng \cite{newell2017pixels} model objects as center points, and ground edges at the midpoint of two vertices then construct the graph via associative embeddings. The midpoint serves as a confidence measurement of presence of relationships. However, false detections and ambiguities arise when there are crowded objects in a region, or the associated objects of a relation are far away from each other. Another limitation is that it still needs feature extraction and grouping that cause low inference speed. Inspired and from a bottom-up 2D human pose estimation work called OpenPose~\cite{cao2019openpose}, we migrate the concept of part affinity fields into scene graph generation. Our proposed method grounds relationships onto CNN features pixel by pixel, and mitigates above mentioned limitations.

\begin{figure}[t]
\centering
\includegraphics[width=0.9\linewidth]{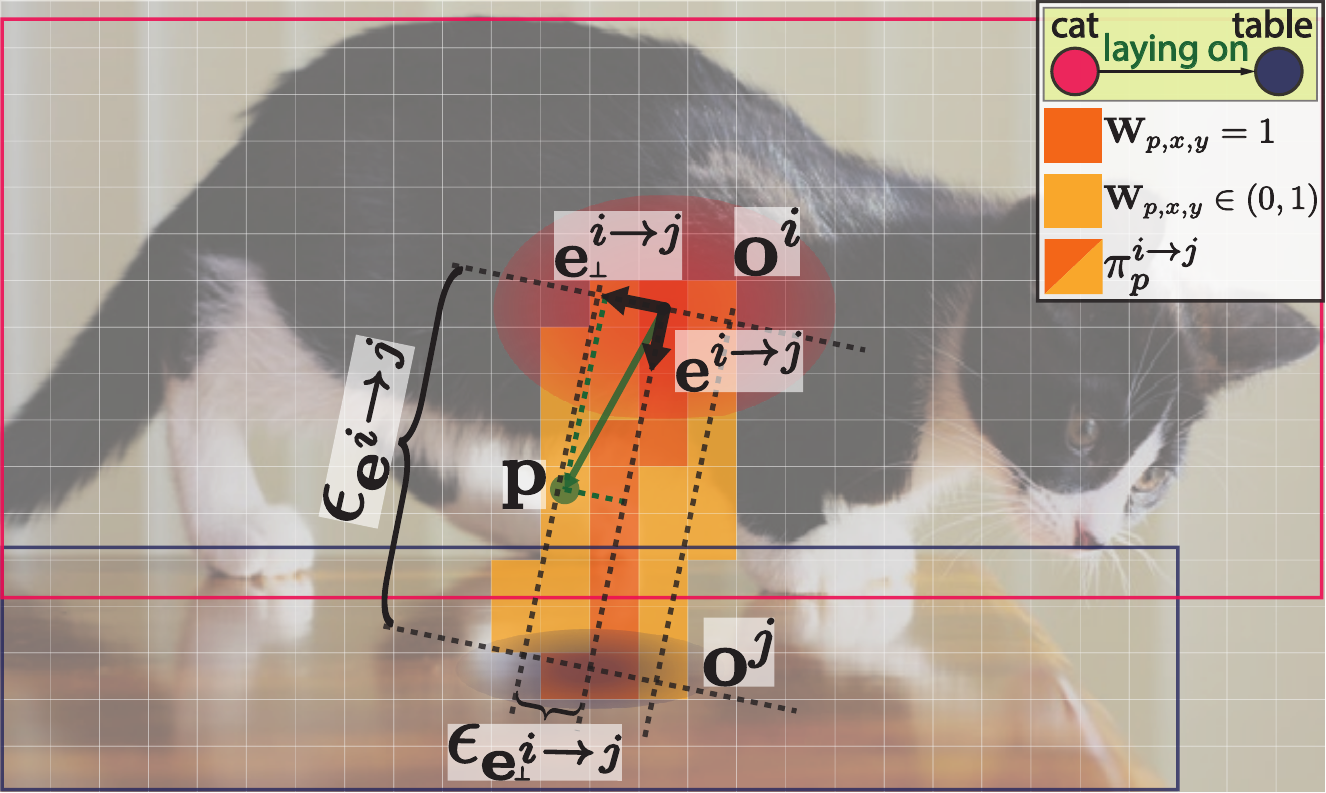}
\caption{An example of GT relation affinity field of predicate \texttt{LAYING ON} based on equations~\ref{eq:raf_vector} and \ref{eq:path}. A non-zero unit vector is only defined on locations inside $\pi^{\texttt{cat} \to \texttt{table}}_{\texttt{LAYING ON}}$. 
}
\label{fig:raf}
\end{figure}

Our model is conceptually simple: in addition to the outputs that are produced by the object detection network described in Section \ref{sec:prelim}, we add another branch that outputs a novel feature representation for relationships called \emph{relation affinity fields} (RAFs). Specifically, the RAFs are a set of 2D vector fields $\mathbf{F} = \{ \mathbf{F}_p \} \in \R^{\mathcal{P} \times 2 \times h \times w}$ , where $p \in \R^{\mathcal{P}} $ and $\mathcal{P}$ is the number of predicate classes in a dataset. Each 2D vector field $\mathbf{F}_p$ represents the relationships among all the object pairs of predicate $p$. Given our definition of objects as center points, the ground-truth RAFs are defined as vectors flow from the center of subject to the center of object. More formally, we define the binary relationships among objects $\mathbf{B}$ in the input image as $\mathbf{R} = \{ r^{i \to j} \}$, where $ r^{i \to j} = ( b^{i}, p^{i \to j}, b^{j} ) $ is the relationship triplet from subject $b^{i}$ to object $b^{j}$ with predicate $p^{i \to j}$. We define a ``path'' $\pi_{p}^ {i \to j}$ that ``propagates'' $p^{i \to j}$ from subject center $\mathbf{o}^{i}$ to object center $\mathbf{o}^{j}$. For a point $\mathbf{p}=(x,y)$, its ground-truth relation affinity field vector $\mathbf{F}_{p,x,y}$ is given as
\begin{equation} \label{eq:raf_vector}
%\resizebox{\linewidth}{!}{$
\mathbf{F}_{p,x,y}=\left\{
\begin{array}{ll}
  \mathbf{e}^{i \to j}=\displaystyle\frac{\mathbf{o}^{j} - \mathbf{o}^{i}}{ || \mathbf{o}^{j} - \mathbf{o}^{i}||_2}   & \text { if } \mathbf{p} \in \pi_p^{i \to j} \\
\mathbf{0} & \text { otherwise, }
\end{array}
\right.
%$}
\end{equation}
and the path $\pi_{p}^ {i \to j}$ is defined on a set of points between object centers forming a rectangular region: 
\begin{equation} \label{eq:path}
\begin{split}
 \pi_{p}^{i \to j} =   \{ \mathbf{p}\, & |\,  0 \leq \mathbf{e}^{i \to j} \cdot (\mathbf{p} -\mathbf{o}^{i}) \leq  \epsilon_{\mathbf{e}^{i \to j}} \\
 & \,\text{ and } 
	|\mathbf{e}_{\perp}^{i \to j} \cdot (\mathbf{p} -\mathbf{o}^{i})| \leq \epsilon_{\mathbf{e}_{\perp}^{i \to j}}
 \}, 
 \end{split}
\end{equation}
where $\epsilon_{\mathbf{e}^{i \to j}} = || \mathbf{o}^{j} - \mathbf{o}^{i}||_2$ as the relationship ``length'' along the direction $\mathbf{e}^{i \to j}$, and $\epsilon_{\mathbf{e}_{\perp}^{i \to j}} = \mathrm{min}(a^{i}, b^{i}, a^{j}, b^{j})$ as the relationship ``semi-width'' along $\mathbf{e}_{\perp}^{i \to j}$ (orthogonal to $\mathbf{e}^{i \to j}$) being the minimum of object centers' radii. Since vectors may overlap at the same point, the ground-truth RAF $\mathbf{F}_{p}$ averages the fields computed for all the relationship triplets containing that particular predicate $p$. It is given as
% \begin{equation} \label{eq:raf}
$\mathbf{F}_{p}= \frac{1}{n_{c}(x,y)} \sum_{x, y} \mathbf{F}_{p,x,y}$, 
% \end{equation}
where $n_{c}(x,y)$ is the number of non-zero vectors at point $(x,y)$. With the definition of ground-truth RAFs, we can train our network to regress such dense feature maps. The RAF regression loss $ \mathcal{L}_{raf} $ can be estimated using a normal regression losses $ \mathcal{L}_{reg} $ such as L1, L2 or smooth L1~\cite{girshick2015fast}. Given the predicted RAFs $\hat{\mathbf{F}}$, the loss is defined as per-pixel weighted regression loss as
\begin{equation} \label{eq:raf_loss}
\mathcal{L}_{raf} = \mathbf{W} \cdot \mathcal{L}_{reg}(\hat{\mathbf{F}}, \mathbf{F}),
\end{equation}
where $\mathbf{W}$ is a pixel-wise weight tensor of the same shape of $\mathbf{F}$. The weights $\mathbf{W}$ are determined and divided into three cases (Figure~\ref{fig:raf}):
\begin {enumerate*} [a) ]%
\item $\mathbf{W}_{p,x,y} = 1$ if $(x,y)$ is exactly on the line segment between objects having the relationship $p$
\item $\mathbf{W}_{p,x,y} \in (0,1)$ if the distance between $(x,y)$ and the line segment is small and the value is negative correlated to the distance 
\item otherwise where $\mathbf{F}_{p,x,y} = \mathbf{0}$. 
\end {enumerate*}
We provide ablation study on the choice of losses and the weight tensor in Section~\ref{sec:ablation}. Finally, the complete loss for training our proposed model can be written as $ \mathcal{L} = \mathcal{L}_{det} + \mathcal{L}_{raf}$.

Our proposed RAFs encode rich information of both spatial and semantic features as dense feature maps, and enable end-to-end joint training of object detection and relation detection. To extract relationship from predictions, path integral over RAFs is performed which will be described below.

\subsection{Inference} \label{sec:inference}

We compute path integrals over RAFs along the line segments connecting pairs of detected object centers as the scores of relationships. Specifically, for two candidate object centers $\hat{\mathbf{o}}^{i}$ and $\hat{\mathbf{o}}^{j}$ with predicted class scores $\hat{h}^{i}$ and $\hat{h}^{j}$, we gather the predicted RAFs $\hat{\mathbf{F}}$ along the path between $\hat{\mathbf{o}}^{i}$ and $\hat{\mathbf{o}}^{j}$, and compute the mean of their projections onto $\hat{\mathbf{e}}^{i \to j}= (\hat{\mathbf{o}}^{j} - \hat{\mathbf{o}}^{i}) /|| \hat{\mathbf{o}}^{j} - \hat{\mathbf{o}}^{i}||_2$. The path integral scores $\mathbf{E}^{i \to j}$ are identified as the confidences of existence of relationships:
\begin{equation}
\mathbf{E}^{i \to j}=
\frac{\hat{h}^{i} \cdot \hat{h}^{j}}{m^{i \to j}} \;
\sum_{p \in \R^{\mathcal{P}}} \sum_{\mathanylap{(x,y) \in \pi^{i \to j}}}			
\mathbf{F}_{p,x,y}
\cdot 
\hat{\mathbf{e}}^{i \to j},
\label{eq:integral}
\end{equation}
where $m^{i \to j} = |\pi^{i \to j}|$ is the number of points in $\pi^{i \to j}$. Since our RAFs are object-class-agnostic, we multiply the class scores of the objects and the path integral score as the overall classification score for the relationship predicate. The integral will be performed spatially for each predicate channel, so $\mathbf{E}^{i \to j}$ represents the confidences of predicted relationship triplet $\hat{r}^{i \to j}$ for all predicates. Note that the integral could be negative that indicates an opposite relationship of the object pairs, and those negative integral values can be simply negated as $\mathbf{E}^{j \to i} = - \mathbf{E}^{i \to j} $. Finally, both scores of $\mathbf{E}^{i \to j}$ and $\mathbf{E}^{j \to i}$ will be used for ranking the predicted relationships. 
We also experiment with a simple re-weighting step known as frequency bias~\cite{zellers2018neural} by multiplying $\mathbf{E}^{i \to j}$ with $1.0001^{n_c(\hat{r}^{i \to j})}$, where $n_c(\hat{r}^{i \to j})$ counts the occurrence of triplet $r^{i \to j}$ in training set. The path integral procedure is presented in Algorithm~\ref{alg:line_integral}. In practice, the operations are performed using matrix multiplication in stead of \textsc{for loop} for fast inference. 

\begin{algorithm}[t]
\begin{algorithmic}[1]
    \renewcommand{\algorithmicrequire}{\textbf{Input:}}
    \renewcommand{\algorithmicensure}{\textbf{Output:}}
    \Require Object centers $\{ \hat{\mathbf{o}}^i \}$, Relation Affinity Fields $\hat{\mathbf{F}}$
    \Ensure Relations $\hat{\mathbf{R}} = \{\hat{r}^{i \to j} = (i, \mathbf{E}^{i \to j}, j)|i \ne j\}$
    \State $\hat{\mathbf{R}} \gets \{\}$ 
%    $ \{ (\mathbf{o}^{i}, \mathbf{o}^{j}) \} \gets $ \Call{combination}{$\{ \hat{\mathbf{o}}^i \}$} $\in \R^{{N\choose{2}} \times 2}$
	\For{each center pair $(\hat{\mathbf{o}}^{i}, \hat{\mathbf{o}}^{j}) |i \neq j$}
	    \State $\pi^{i \to j} \gets $ \Call{linspace}{$\hat{\mathbf{o}}^{i}, \hat{\mathbf{o}}^{j}$} $ \in \R^{m^{i \to j} \times 2}$
	    \State $\hat{\mathbf{F}}_{\pi^{i \to j}}$ $\gets$ \Call{index}{$\hat{\mathbf{F}}$, $\pi^{i \to j}$} $\in \R^{\mathcal{P} \times m^{i \to j} \times 2}$
	    \State $\mathbf{E}^{i \to j}$, $\mathbf{E}^{j \to i} \in \R^{\mathcal{P}}$ $\gets$ Equation \ref{eq:integral}
	    \State{$
	    	\hat{\mathbf{R}} \gets \hat{\mathbf{R}}
                  \bigcup \{ \hat{r}^{i \to j} = (i, \mathbf{E}^{i \to j}, j) \} 
              $}
        \State{$
                 \qquad\quad\,\,\, \bigcup \{ \hat{r}^{j \to i} = (j, \mathbf{E}^{j \to i}, i) \}
        	  $}
	\EndFor
	\State \Return \Call{sorted}{$\hat{\mathbf{R}}$}
% 	\EndFunction
\end{algorithmic}
\caption{Path Integral over Relation Affinity Fields}
\label{alg:line_integral}
\end{algorithm}

\subsection{Multi-scale Prediction} \label{sec:ms}

Since object centers are downsampled to feature level, their centers could collide onto the same pixel location. Regression ambiguity may rise due to single-scale feature representations. In this subsection, we address this problem by utilizing multi-scale prediction and shared detection heads.

%The reason is that object centers will be assigned to individual channel of $\mathbf{O}$ based on object class, so that even two objects share the same center location, it is unlikely to for them to be the same class. For object-agnostic size and offset regression, the GT location is tied to center location thus the regression ambiguity occurs.
 Though Zhou~\etal~\cite{zhou2019objects} argued that only a very small fraction (\textless 0.1\% in COCO~\cite{lin2014microsoft} dataset) of objects have center collision problem at stride of 4, the size and offset regression targets need better assignment strategy since there is only one valid target per pixel. We follow the work of FPN~\cite{lin2017feature}, RetinaNet~\cite{lin2017focal} and FCOS~\cite{tian2019fcos}, and assign the ground-truth bounding boxes to different levels based on scales. Building upon the backbone features, we construct multi-level feature maps $\{\text{P}_k\}$ where $\text{P}_k$ is of stride $2^k$. We refer the network component of generating multi-scale features as the ``neck'' (the green box in Figure~\ref{fig:architecture}), such as FPN~\cite{lin2017feature}. We define a valid range $[l_k, u_k] \subset [0, \text{L}_{max}] $ for objects in each scale, where $\text{L}_{max}$ is the maximum size of longer edge allowed for training and testing. Only bounding boxes of area within $[l_k^2, u_k^2]$ are qualified for the k-th scale training. We experiment different number of scale levels and input image size. For smaller input image of $\text{L}_{max}$ = 512, we build 4-scale features~\cite{lin2017feature} $\{ \text{P}_2, \text{P}_3, \text{P}_4, \text{P}_5\}$ (as shown in Figure~\ref{fig:architecture}) with valid ranges \{[0, 32$^2$], [32$^2$, 64$^2$], [64$^2$, 128$^2$], [128$^2$, 512$^2$]\}; for larger input image of $\text{L}_{max}$ = 1024 (shorter edge is at least $\text{L}_{min}$ = 640), we use 5-scale features~\cite{lin2017focal, tian2019fcos} $\{ \text{P}_3, \text{P}_4, \text{P}_5, \text{P}_6, \text{P}_7\}$ with area ranges \{[0, 64$^2$], [64$^2$, 128$^2$], [128$^2$, 256$^2$], [256$^2$, 512$^2$], [512$^2$, 1024$^2$]\}. If there is still more than one target at the same location, we simply choose the smallest object for regression. 

In terms of the multi-scale RAFs training, the GT assignment is based on the distances between object centers. For high-level semantic features like P$_5$, the feature map can capture large objects, as well as relationships among distant objects. We select the relationships of ``length'' $\epsilon_{\mathbf{e}^{i \to j}} \in [l_k, u_k]$ as valid samples for training the k-th scale. 
The exact ranges for 4-scale or 5-scale RAFs are the same as the settings for bounding boxes.
Finally, the weights of detection heads are shared across different feature scales for efficiencies and performance improvements. During inference, we gather outputs from each scale based on the corresponding valid range, then merge and rank all relationship triplets. 
Figure~\ref{fig:architecture} illustrates the details of our proposed architecture using a four-scale feature setup as an example with shared detection heads.
Our experiments (Section~\ref{sec:experiments}) show that the multi-scale GT and scale-aware training resolve the aforementioned ambiguity problem thus improve the results over single-scale prediction.

%------------------------------------------------------------------------

\section{Experiments} \label{sec:experiments}

\newcommand\RotText[1]{\rotatebox{90}{\parbox{2cm}{\centering#1}}}
\newcolumntype{A}{m{0.0em}}
\newcolumntype{B}{w{r}{2em}}
\newcolumntype{C}{w{r}{2em}}
\newcolumntype{D}{@{\hspace{0.1em}}c@{\hspace{0.1em}}}
\newcolumntype{G}{m{0.2em}}

\begin{table*}[t]
\setlength\tabcolsep{3pt}
\resizebox{\linewidth}{!}{%
\begin{tabular}{c|l|c|CDBDB A CDBDB G|CDBDB A CDBDB G|CDBDB A CDBDB G} % 39 cols
\hline 
 \multicolumn{2}{l|} {Recall @K /} & \multirow{2}{*}{AP$_{50}$} & \multicolumn{12}{c|} {Predicate Classification} & \multicolumn{12}{c|} { Scene Graph Classification } & \multicolumn{12}{c} { Scene Graph Detection } \\ 
 
\multicolumn{2}{l|}{No-graph Constraint Recall @K} & & \multicolumn{5}{c}{R@20/50 /100} & & \multicolumn{6}{c|}{ng-R@20/50/100} & \multicolumn{5}{c}{R@20/50/100} & \multicolumn{1}{c}{} & \multicolumn{6}{c|}{ng-R@20/50/100} & \multicolumn{5}{c}{R@20/50/100} & \multicolumn{1}{c}{} & \multicolumn{6}{c}{ng-R@20/50/100} \\\hline
 
 \multirow{3}{*}{\RotText{External Knowledge}} & VCTree~\cite{tang2019learning} & - & 60.1 & / & 66.4 & / & 68.1 & & \multicolumn{5}{c}{-} & \multicolumn{1}{c|}{} & 35.2 & / & 38.1 & / & 38.8 &  & \multicolumn{5}{c}{-} & \multicolumn{1}{c|}{} & 22.0 & / & 27.9 & / & 31.3 & & \multicolumn{5}{c}{-} & \multicolumn{1}{c}{}\\
& KERN~\cite{chen2019knowledge} & - & - & / & 65.8 & / & 67.6 & & - & / & 81.9 & / & 88.9 & & - & / & 36.7 & / & 37.4 & & - & / & 45.9 & / & 49.0 & & - & / & 27.1 & / & 29.8 &  & - & / & 30.9 & / & 35.8 & \\
& GPS-Net~\cite{lin2020gps} & - & 67.6 & / & 69.7 & / & 69.7 & & \multicolumn{5}{c}{-} & \multicolumn{1}{c|}{} & 41.8 & / & 42.3 & / & 42.3 & & \multicolumn{5}{c}{-} & \multicolumn{1}{c|}{} & 22.3 & / & 28.9 & / & 33.2 & & \multicolumn{5}{c}{-} & \multicolumn{1}{c}{}\\
& MOTIFS-TDE~\cite{tang2020unbiased,zellers2018neural} & 28.1 & 33.6 & / & 46.2 & / & 51.4 & & \multicolumn{5}{c}{-} & \multicolumn{1}{c|}{} & 21.7 & / & 27.7 & / & 29.9 & & \multicolumn{5}{c}{-} & \multicolumn{1}{c|}{} & 12.4 & / & 16.9 & / & 20.3 & & \multicolumn{5}{c}{-} & \multicolumn{1}{c}{}\\
& GB-N{\footnotesize ET}-$\beta$~\cite{zareian2020bridging} & - & - & / & 66.6 & / & 68.2 & & - & / & 83.5 & / & 90.3 & \multicolumn{1}{c|}{} & - & / & 37.3 & / & 38.0 & & - & / & 46.9 & / & 50.3 & \multicolumn{1}{c|}{} & - & / & 26.3 & / & 29.9 & & - & / & 29.3 & / & 35.0 & \multicolumn{1}{c}{} \\
\hline
\multirow{7}{*}{\RotText{Visual Only}} & VTransE$^\star$ \cite{zhang2017visual} & - & \multicolumn{5}{c}{-} & & \multicolumn{5}{c}{-} & \multicolumn{1}{c|}{} & \multicolumn{5}{c}{-} & & \multicolumn{5}{c}{-} & \multicolumn{1}{c|}{} & - & / & 5.5 & / & 6.0 & & \multicolumn{5}{c}{-} & \multicolumn{1}{c}{} \\
& FactorizableNet$^{\star}$~\cite{li2018factorizable} & - & \multicolumn{5}{c}{-} & & \multicolumn{5}{c}{-} & \multicolumn{1}{c|}{} & \multicolumn{5}{c}{-} & & \multicolumn{5}{c}{-} & \multicolumn{1}{c|}{} & - & / & 13.1 & / & 16.5 & & \multicolumn{5}{c}{-} & \multicolumn{1}{c}{} \\	
% & IMP \cite{xu2017scene} & - & -\hspace{.58em} / 44.8 / 53.1 & - & \hspace{.8em}-\hspace{.58em} / 21.7 / 24.4 & - & \hspace{.8em}-\hspace{.58em} / 3.4 / 4.2 & - \\
& IMP$^\dagger$ \cite{xu2017scene,zellers2018neural} & 20.0 & 58.5 & / & 65.2 & / & 67.1 & & \multicolumn{5}{c}{-} & \multicolumn{1}{c|}{} & 31.7 & / & 34.6 & / & 35.4 & & \multicolumn{5}{c}{-} & \multicolumn{1}{c|}{} & 14.6 & / & 20.7 & / & 24.5 & & \multicolumn{5}{c}{-} & \multicolumn{1}{c}{}\\
% Pixels2Graphs \cite{newell2017pixels} & - & - & \hspace{.8em}-\hspace{.58em} / 68.0 / 75.2 & - & \hspace{.8em}-\hspace{.58em} / 26.5 / 30.0 & - & \hspace{.8em}-\hspace{.58em} / 9.7 / 11.3 \\
& Pixels2Graphs \cite{newell2017pixels} & - & \multicolumn{5}{c}{-} & & - & / & 68.0 & / & 75.2 & \multicolumn{1}{c|}{} & \multicolumn{5}{c}{-} & & - & / & 26.5 & / & 30.0 & \multicolumn{1}{c|}{} & \multicolumn{5}{c}{-} & & - & / & 9.7 & / & 11.3 & \multicolumn{1}{c}{}\\
% Graph R-CNN \cite{yang2018graph} & 23.0 & \hspace{.8em}-\hspace{.58em} / 54.2 / 59.1 & - & \hspace{.8em}-\hspace{.58em} / 29.6 / 31.6 & - & \hspace{.8em}-\hspace{.58em} / 11.4 / 13.7 & - \\
& Graph R-CNN \cite{yang2018graph} & 23.0 & - & / & 54.2 & / & 59.1 & & \multicolumn{5}{c}{-} & \multicolumn{1}{c|}{} & - & / & 29.6 & / & 31.6 & & \multicolumn{5}{c}{-} & \multicolumn{1}{c|}{} & - & / & 11.4 & / & 13.7 & & \multicolumn{5}{c}{-} & \multicolumn{1}{c}{}\\
% VRF~\cite{dornadula2019visual} & - & \hspace{.8em}-\hspace{.58em} / 56.7 / 57.2 & - & \hspace{.8em}-\hspace{.58em} / 23.7 / 24.7 & - & \hspace{.8em}-\hspace{.58em} / 13.2 / 13.5 & - \\ 
& VRF~\cite{dornadula2019visual} & - & - & / & 56.7 & / & 57.2 & & \multicolumn{5}{c}{-} & \multicolumn{1}{c|}{} & - & / & 23.7 & / & 24.7 & & \multicolumn{5}{c}{-} & \multicolumn{1}{c|}{} & - & / & 13.2 & / & 13.5 & & \multicolumn{5}{c}{-} & \multicolumn{1}{c}{}\\
% CISC~\cite{wang2019exploring} & - & 42.1 / 53.2 / 57.9 & - & 23.3 / 27.8 /29.5 & - & \hspace{.4em}7.7\hspace{.1em} / 11.4 / 13.9 & - \\
& CISC~\cite{wang2019exploring} & - & 42.1 & / & 53.2 & / & 57.9 & & \multicolumn{5}{c}{-} & \multicolumn{1}{c|}{} & 23.3 & / & 27.8 & / & 29.5 & & \multicolumn{5}{c}{-} & \multicolumn{1}{c|}{} & 7.7 & / & 11.4 & / & 13.9 & & \multicolumn{5}{c}{-} & \multicolumn{1}{c}{}\\
\hline
% \multicolumn{7}{l}{\textbf{FCSGG} (\textbf{Ours})} \\\hline
\multirow{7}{*}{\vspace{2em}\RotText{\textbf{FCSGG} (\textbf{Ours})}} & HRNetW32-1S & 21.6 & 27.6 & / & 34.9 & / & 38.5 & & 32.2 & / & 46.3 & / & 56.6 & \multicolumn{1}{c|}{} & 12.3 & / & 15.5 & / & 17.2 & & 13.5 & / & 19.3 & / & 23.6 & \multicolumn{1}{c|}{} & 11.0 & / & 15.1 & / & 18.1 & & 12.4 & / & 18.2 & / & 23.0 & \multicolumn{1}{c}{}\\
& HRNetW48-1S & 25.0 & 24.2 & / & 31.0 & / & 34.6 & & 28.1 & / & 40.3 & / & 50.0 & \multicolumn{1}{c|}{} & 13.6 & / & 17.1 & / & 18.8 & & 14.2 & / & 19.6 & / & 24.0 & \multicolumn{1}{c|}{} & 11.5 & / & 15.5 & / & 18.4 & & 12.7 & / & 18.3 & / & 23.0 & \multicolumn{1}{c}{}\\

& ResNet50-4S-FPN{\scriptsize$\times 2$} & 23.0 & 28.0 & / & 35.8 & / & 40.2 & & 31.6 & / & 44.7 & / & 54.8 & \multicolumn{1}{c|}{} & 13.9 & / & 17.7 & / & 19.6 & & 14.8 & / & 20.6 & / & 25.0 & \multicolumn{1}{c|}{} & 11.4 & / & 15.7 & / & 19.0 & &  12.2 & / & 18.0 & / & 22.8 & \multicolumn{1}{c}{}\\

& HRNetW48-5S-FPN{\scriptsize$\times 2$} & 28.5 & 28.9 & / & 37.1 & / & 41.3 & & 34.0 & / & 48.1 & /  & 58.4 & \multicolumn{1}{c|}{} & 16.9 & / & 21.4 & / & 23.6 & & 18.6 & / & 26.1 & / & 31.6 & \multicolumn{1}{c|}{} & 13.5 & / & 18.4 & / & 22.0 & &  15.4 & / & 22.5 & / & 28.3 & \multicolumn{1}{c}{}\\

& HRNetW48-5S-FPN{\scriptsize$\times 2$}-\textit{f} & 28.5 & 33.4 & / & 41.0 & / & 45.0 & & 37.2 & / & 50.0 & / & 59.2 & \multicolumn{1}{c|}{} & 19.0 & / & 23.5 & / & 25.7 & & 19.6 & / & 26.8 & / & 32.1 & \multicolumn{1}{c|}{} & 16.1 & / & 21.3 & / & 25.1 & & 16.7 & / & 23.5 & / & 29.2 & \multicolumn{1}{c}{} \\
\hline
\end{tabular}
}

\caption{Recall and no-graph constraint recall @K evaluation results on VG-150.  $\star$ denotes the methods evaluated on other datasets, such that VTransE is evaluated on VG-200 \cite{zhang2017visual} and FactorizableNet on a smaller set following \cite{li2017scene}. $\dagger$ denotes the methods with updated re-implementation results. - denotes the results that are not reported in the corresponding work.}
\label{tab:recall}
\end{table*}

\newcolumntype{E}{w{r}{2em}}
\newcolumntype{F}{w{r}{1.8em}}
\newcolumntype{H}{m{0.4em}}
\newcolumntype{I}{@{\hspace{0.2em}}c@{\hspace{0.2em}}}
\newcolumntype{J}{m{0.01em}}
\newcolumntype{K}{@{\hspace{0.1em}}c@{\hspace{0.1em}}}
\newcolumntype{L}{m{0.3em}}
\newcolumntype{M}{@{\hspace{-0.2em}}w{r}{1em}}
\begin{table*}[t]
\setlength\tabcolsep{3pt}
\centering
\resizebox{1.0\linewidth}{!}{%
\begin{tabular}{l|FIFIF L FIFIF J|FIFIF L FIFIF J|FIFIF L FIFIF J} % 37 cols
\hline 
 Mean Recall @K / & \multicolumn{12}{c|} { Predicate Classification } & \multicolumn{12}{c|} { Scene Graph Classification } & \multicolumn{12}{c} { Scene Graph Detection } \\
Ng Mean Recall @K & \multicolumn{6}{c} {mR@20/50/100} & 
\multicolumn{6}{c|} {ng-mR@20/50/100} & \multicolumn{6}{c} {mR@20/50/100} & 
\multicolumn{6}{c|} {ng-mR@20/50/100} & \multicolumn{6}{c} {mR@20/50/100} & 
\multicolumn{6}{c} {ng-mR@20/50/100} \\\hline

VCTree \cite{tang2019learning} & 
14.0 & / & 17.9 & / & 19.4 & & \multicolumn{6}{c|}{-} & 8.2 & / & 10.1 & / & 11.8 & & \multicolumn{6}{c|}{-}
& 5.2 & / & 6.9 & / & 8.0 & & \multicolumn{6}{c}{-}\\
KERN \cite{chen2019knowledge} &
- & / & 17.7 & / & 19.4 & & \multicolumn{6}{c|}{-} & 
- & / & 9.4 & / & 10.0 & & \multicolumn{6}{c|}{-} & 
- & / & 6.4 & / & 7.3 & & \multicolumn{6}{c}{-}\\
GPS-Net~\cite{lin2020gps}  &
- & / & - & / & 22.8 & & \multicolumn{6}{c|}{-} & 
- & / & - & / & 12.6 & & \multicolumn{6}{c|}{-} & 
- & / & - & / & 9.8 & & \multicolumn{6}{c}{-}\\
MOTIFS-TDE \cite{tang2020unbiased,zellers2018neural}  &
18.5 & / & 25.5 & / & 29.1 & & \multicolumn{6}{c|}{-} & 
9.8 & / & 13.1 & / & 14.9 & & \multicolumn{6}{c|}{-} & 
5.8 & / & 8.2 & / & 9.8 & & \multicolumn{6}{c}{-}\\
GB-N{\footnotesize ET}-$\beta$~\cite{zareian2020bridging} &
- & / & 22.1 & / & 24.0 & & \multicolumn{6}{c|}{-} & 
- & / & 12.7 & / & 13.4 & & \multicolumn{6}{c|}{-} & 
- & / & 7.1 & / & 8.5 & & \multicolumn{6}{c}{-}\\\hline
HRNetW32-1S & 4.0 & / & 5.5 & / & 6.3 & & 5.4 & / & 9.7 & / & 13.6 & & 1.9 & / & 2.5 & / & 2.8 & & 2.7 & / & 4.4 & / & 6.2 & & 1.7 & / & 2.4 & / & 2.9 & & 2.2 & / & 3.6 & / & 4.9 & \\
HRNetW48-1S & 3.7 & / & 5.2 & / & 6.1 & & 5.2 & / & 9.5 & / & 14.7 & & 2.2 & / & 2.9 & / & 3.4 & & 3.5 & / & 6.3 & / & 9.4 & & 1.8 & / & 2.6 & / & 3.1 & & 2.7 & / & 4.7 & / & 6.9 & \\
ResNet50-4S-FPN{\scriptsize$\times 2$} & 4.2 & / & 5.7 & / & 6.7 & & 6.5 & / & 11.3 & / & 16.6 & & 2.2 & / & 2.9 & / & 3.3 & & 3.6 & / & 6.0 & / & 8.3 & & 1.9 & / & 2.7 & / & 3.3 & & 3.0 & / & 4.9 & / & 6.8 & \\
HRNetW48-5S-FPN{\scriptsize$\times 2$} & 4.3 & / & 5.8 & / & 6.7 & & 6.1 & / & 10.3 & / & 14.2 & & 2.6 & / & 3.4 & / & 3.8 & & 4.1 & / & 6.4 & / & 8.4 & & 2.3 & / & 3.2 & / & 3.8 & & 3.7 & / & 5.7 & / & 7.4 & \\
HRNetW48-5S-FPN{\scriptsize$\times 2$}-\textit{f} & 4.9 & / & 6.3 & / & 7.1 & & 6.6 & / & 10.5 & / & 14.3 & & 2.9 & / & 3.7 & / & 4.1 & & 4.2 & / & 6.5 & / & 8.6 & & 2.7 & / & 3.6 & / & 4.2 & & 3.8 & / & 5.7 & / & 7.5 & \\

\hline
\end{tabular}
}
\caption{The SGG results on mean recall@K and no-graph constraint mean recall@K.}
\label{tab:mrecall}
\end{table*}

\emph{Dataset}. We use the Visual Genome (VG) \cite{krishna2017visual} dataset to train and evaluate our models. We followed the widely-used preprocessed subset of VG-150 \cite{xu2017scene} which contains the most frequent 150 object categories ($\mathcal{C}$ =   150) and 50 predicate categories ($\mathcal{P}$ = 50). The dataset contains approximately 108k images, with 70\% for training and 30\% for testing. Different from previous works~\cite{chen2019counterfactual,tang2019learning,zellers2018neural}, we do not filter non-overlapping triplets for evaluation. 
%We also filter out duplicated objects and relationship triplets during training, and note that we did not remove those of same subject-object pair with multiple different relations as our RAFs naturally support multi-label cases.

\emph{General settings}. 
We experiment on two settings, one for small input size ($\text{L}_{max}$ = 512) and one for larger size ($\text{L}_{max}$ = 1024). 
% For the former, input images are resized such that their longer edge is no larger than 512 pixels. For the latter, images are resized such that their shorter edge is at least 640 pixels and their longer edge is no larger than 1024 pixels. 
The model is trained end-to-end using SGD optimizer with the batch size of 16 for 120k iterations. The initial learning rate is set to 0.02 and decayed by the factor of 10 at 80k$^{\text{th}}$ and 100k$^{\text{th}}$ iteration. We adopt standard image augmentations of horizontal flip and random crop with multi-scale training. During testing, we keep the top 100 detected objects for path integral.

\emph{Metrics}. 
We conduct comprehensive analysis following three standard evaluation tasks: Predicate Classification (PredCls), Scene Graph Classification (SGCls), and Scene Graph Detection (SGDet).
We report results of recall@K (R@K)~\cite{lu2016visual}, no-graph constraint recall@K (ng\-R@K)~\cite{newell2017pixels,zellers2018neural}, mean recall@K (mR@K)~\cite{chen2019knowledge,tang2019learning}, no-graph constraint mean recall@K (ng-mR@K), zero-shot recall@K (zsR@K)~\cite{lu2016visual} and no-graph constraint zero-shot recall@K (ng-zsR@K)~\cite{tang2020unbiased} for all three evaluation tasks. We do not train separate models for different tasks.

\subsection{Implementation Details}

We conduct experiments on different backbone and neck networks. Each of the detection heads consists of four $3 \times 3$ convolutions followed by batch normalization and ReLU, and one $1 \times 1$ convolution with the desired number of output channels in all our experiments unless specified otherwise. For convenience, our models are named as \textsc{backbone\,-\,\# of output scales\,-\,neck\,-\,other options}. 
% We summarize a model zoo for results comparisons:

\emph{ResNet}~\cite{he2016deep, lin2017feature}. 
We start by using ResNet-50 as our backbone and build a 4-scale FPN for multi-scale prediction. Since the tasks of object detection and RAFs prediction are jointly trained, the losses from the two tasks could compete with each other. We implement a neck named ``FPN{\scriptsize$\times 2$}'' with two parallel FPNs, such that one FPN is used for constructing features for object detection heads (center, size and offset), and the other is for producing features for RAFs. We name this model as ResNet50-4S-FPN{\scriptsize$\times 2$}.

\emph{HRNet}~\cite{wang2020deep}. 
We then experiment on a recent proposed backbone network called HRNet that consists of parallel convolution branches with information exchange across different scales. For single-scale experiments, we use HRNetV2-W32 and HRNetV2-W48; and for multi-scale prediction, we adopt its pyramid version called HRNetV2p. We omit their version number for the rest of the paper. We test several models: HRNetW32-1S, HRNetW48-1S and HRNetW48-5S-FPN{\scriptsize$\times 2$}. We also experiment on adding frequency bias for inference as discussed in Section~\ref{sec:inference}, and the model used is called HRNetW48-5S-FPN{\scriptsize$\times 2$}-\textit{f}.

\subsection{Quantitative Results}

\begin{figure*}[t]
% \centering
\begin{minipage}[b]{0.615\textwidth}
\centering
\captionsetup{type=table} %% tell latex to change to table
% \begin{table}[t]
% \centering
\resizebox{\textwidth}{!}{%
\begin{tabular}{l|cc|cc|cc}
\hline 
 \multicolumn{1}{l|} { Zero-shot Recall @K } & \multicolumn{2}{c|} { PredCls } & \multicolumn{2}{c|} { SGCls } & \multicolumn{2}{c} { SGDet } \\
 \cline{1-1} Method & \multicolumn{2}{c|} {zsR@50/100} 
 & \multicolumn{2}{c|} {zsR@50/100}
  & \multicolumn{2}{c} {zsR@ 50/100} \\\hline 
MOTIFS-TDE \cite{tang2020unbiased} & \multicolumn{2}{c|}{14.4 / 18.2}
 & \multicolumn{2}{c|}{3.4 / 4.5}
  & \multicolumn{2}{c}{2.3 / 2.9} \\
VTransE-TDE \cite{tang2020unbiased} 
& \multicolumn{2}{c|}{13.3 / 17.6} 
& \multicolumn{2}{c|}{2.9 / 3.8} 
& \multicolumn{2}{c}{2.0 / 2.7}  \\
VCTree-TDE \cite{tang2020unbiased} 
& \multicolumn{2}{c|}{14.3 / 17.6}
& \multicolumn{2}{c|}{3.2 / 4.0} 
& \multicolumn{2}{c}{2.6 / 3.2}  \\
Knyazev~\etal \cite{knyazev2020graph} 
& \multicolumn{2}{c|}{\hspace{0.8em}-\hspace{0.65em} / 21.5} 
& \multicolumn{2}{c|}{\hspace{0.5em}-\hspace{0.5em} / 4.2} 
& \multicolumn{2}{c}{\hspace{0.1em}-\hspace{0.4em}/ \hspace{0.1em}-\hspace{0.5em}}  \\
\hline 
\multirow{2}{*}{\textbf{FCSGG} (\textbf{Ours})} 
& zsR & ng-zsR & zsR & ng-zsR & zsR & ng-zsR \\ 
%\multicolumn{1}{l|}{} 
& @50/100 & @50/100 & @50/100 & @50/100 & @50/100 & @50/100 \\ \hline
HRNetW32-1S & 8.3 / 10.7 & 12.9 / 19.2 & 1.0 / 1.2 & 2.3 / 3.5 & 0.6 / 1.0 & 1.2 / 1.6 \\
HRNetW48-1S & 8.6 / 10.9 & 12.8 / 19.6 & 1.7 / 2.1 & 2.9 / 4.4 & 1.0 / 1.4 & 1.8 / 2.7  \\
%ResNet50-4S-FPN & & - & -  \\
ResNet50-4S-FPN{\scriptsize$\times 2$}  & 8.2 / 10.6 & 11.7 / 18.1 & 1.3 / 1.7 & 2.4 / 3.8 & 0.8 / 1.1 & 1.0 / 1.7 \\
%HRNetW32-4S-FPN{\scriptsize$\times 2$} & - & - & -   \\
HRNetW48-5S-FPN{\scriptsize$\times 2$} & 7.9 / 10.1 & 11.5 / 17.7 & 1.7 / 2.1 & 2.8 / 4.8 & 0.9 / 1.4 & 1.4 / 2.4   \\
HRNetW48-5S-FPN{\scriptsize$\times 2$}-\textit{f} & 7.8 / 10.0 & 11.4 / 17.6 & 1.6 / 2.0 & 2.8 / 4.8 & 0.8 / 1.4 & 1.4 / 2.3 \\
\hline

\end{tabular}
}
\caption{Comparisons of SGG results on zero-shot Recall@K, and our results on no-graph constraint zero-shot Recall@K.}
\label{tab:zsr}
% \end{table}
\end{minipage}%
\hfill
\begin{minipage}[b]{0.345\textwidth}
\centering
% \begin{figure}[t]
% \centering
\includegraphics[width=\linewidth]{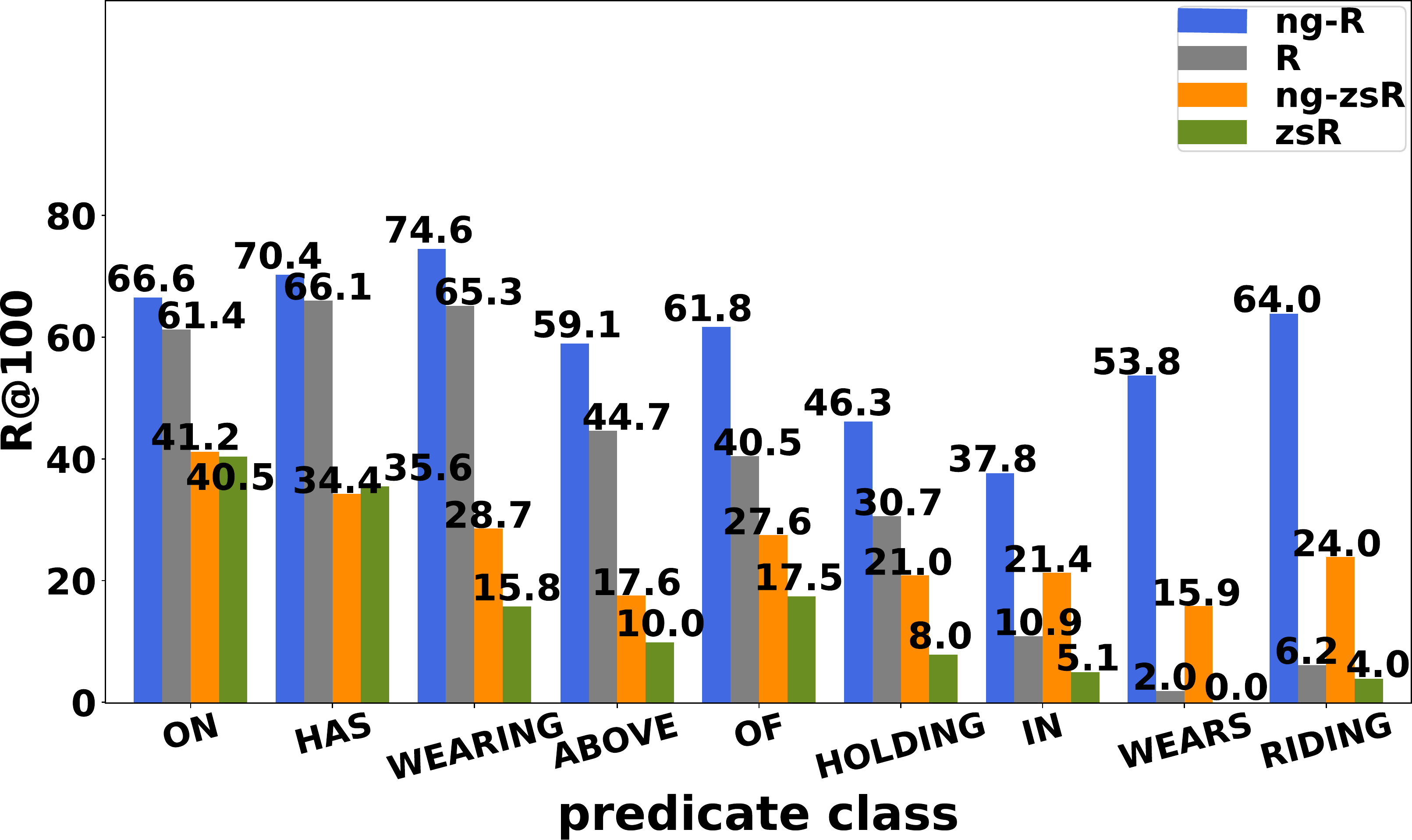}
\caption{FCSGG per-predicate PredCls@100 results for selected predicates using HRNetW48-5S-FPN{\scriptsize$\times 2$}-\textit{f}.
}
\label{fig:perclass}
% \end{figure}
\end{minipage}
\end{figure*}

We first compare results of R@K and ng-R@K with various of state-of-the-art (SOTA) models, and divide them into two categories: 
\begin {enumerate*} [1) ]%
\item models that only use visual features derived from the input image like our proposed model
\item models that not only use visual features, but also use features like language embeddings, dataset statistics or counterfactual causality, \etc
\end {enumerate*} 
 The results are shown in Table~\ref{tab:recall}. Our best model achieves 28.5 average precision at IoU = 0.5 ($\mathrm{AP_{50}}$) for object detection. Though our SGG results do not outperform the SOTA approaches, we achieve the best scene graph detection results among visual-only models. Specifically, Pixels2Graphs~\cite{newell2017pixels} and our models are the only models without using Faster R-CNN~\cite{ren2015faster} as object detector, and we achieves 13.8 / 17.9 gain on SGDet ng-R@50 / 100 compared with their results using RPN~\cite{ren2015faster}. We then report mean recall and no-graph constraint mean recall results shown in Table~\ref{tab:mrecall}. We still obtain competitive results, especially on ng-mR. By comparing (m)\,R@K and ng-(m)\,R@K directly, we observe more gain on our results than other methods'. 
 
Zero-shot recall (zsR)~\cite{gkanatsiossaturation,hung2020contextual,knyazev2020graph,lu2016visual,plummer2017phrase,tang2020unbiased,yang2018shuffle} is a proper metric for evaluating the model's robustness and generalizability for generating scene graphs. It computes recall on those subject-predicate-object triplets that do not present during training. There are in total of 5971 unique zero-shot triplets from the testing set of VG-150. The results and comparisons are listed in Table~\ref{tab:zsr}. We also compute the per-predicate recall@100 for predicate classification task using HRNetW48-5S-FPN{\scriptsize$\times 2$}-\textit{f}, and show the comparisons in Figure~\ref{fig:perclass}. We observe similar behavior with our results on recall, such that the results on no-graph constraint zero-shot recall are significantly better than zero-shot recall. Even for unseen triplets, purely based on visual features, FCSGG is still capable of predicting meaningful RAFs which proves its generalization capability. In other words, when constructing scene graphs from RAFs, our approach does not highly depend on the object classes but only focuses on the context features between the objects. When comparing with other reported results on zsR, we achieve slightly lower results than those much larger models. For example of PredCls task, ResNet50-4S-FPN{\scriptsize$\times 2$} achieves 10.6 zsR@100 with only 36 million (M) number of parameters and inference time of 40 milliseconds (ms) per image, while VCTree-TDE~\cite{tang2020unbiased} achieves 17.6 zsR@100 with 360.8M number of parameters and inference time of 1.69\,s per image. For comparison, ResNet50-4S-FPN{\scriptsize$\times 2$} is \emph{10 times} smaller and \emph{42 times} faster than VCTree-TDE.
 
% The figure shows that our model achieves good performances for detecting zero-shot relationships (zsR and ng-zsR) as well as other ``seen'' relationships (R and ng-R) which further demonstrates the generalizability of our proposed RAFs.
% 
% It is worth noting that for all SGG evaluation metrics, there are trade-offs among recall, mean recall and zero-shot recall since the top-K ranking is task-agnostic. (Tang~\etal~\cite{tang2020unbiased} achieve high mR by incorporating total direct effect (TDE), but the recall significantly drops as a result.) This trade-off has not been observed before for zero-shot setting since the zsR results are always poor for previous methods. Among the top-K relationships, we detect more valid ones from unseen triplets resulting in the performance drop on recall and mean recall. 
 \emph{Limitations}. It should be noticed that FCSGG also has some ``disadvantages'' over Faster R-CNN-based methods on easier tasks such as PredCls and SGCls. During evaluations with given GT bounding boxes or classes, our RAFs features will not change, while R-CNN extracted object/union-box features will change which leads to better results. When using visual-only representation of relationships, it is hard for the network to distinguish predicates between \texttt{WEARS} / \texttt{WEARING} (by comparing R and ng-R in Figure~\ref{fig:perclass}) or \texttt{LAYING ON} / \texttt{LYING ON}, which is common in VG dataset. In this sense, incorporating external knowledge gives FCSGG a large potential in improving results. Comparing the model HRNetW48-5S-FPN{\scriptsize$\times 2$} and its frequency-biased counterpart HRNetW48-5S-FPN{\scriptsize$\times 2$}-\textit{f}, we find noticeable improvement by using training set statistics. This simple cost-free operation can improve R@20 by 2.6, and we expect better results from fine-tuning hyper-parameters. However, the focus of this work is not perfectly fitting on a dataset, but improving generalization of relationship based on visual features. More sophisticated ensemble methods or extensions are beyond the scope of this paper. 

\subsection{Ablation Study} \label{sec:ablation}
\subsubsection{RAF regression Loss}

We experience the same difficulty of training from sparsely annotated scene graphs as discussed by Newell~\etal~\cite{newell2017pixels}. The network has the potential of generating reasonable triplets not covered in the ground-truth, and our results on zero-shot recall prove the argument. To reduce the penalty on these detections, we investigate the design methodology of RAF regression loss $\mathcal{L}_{raf}$ (Equation~\ref{eq:raf_loss}). 

\begin{table}[b]
\centering
\resizebox{0.8\columnwidth}{!}{%
\begin{tabular}{r|r|c|Z{2.3em} Z{2.3em} Z{2.3em}}
$\mathcal{L}_{raf}^{+}$ & $\beta$ & $\mathrm{AP}_{50}$ & R@50 & zR@50 & mR@50 \\
\hline 
L1 & 0 & 21.57 & 6.22 & 0.40 & 2.28 \\
L1 & 1 & 21.52 & 9.80 & 0.56 & 2.33\\ 
L1 & 10 & 21.56 & 15.05 & 0.60 & 2.36\\ \hline
Smooth L1 & 0 & 20.15 & 5.00 & 0.30 & 2.51 \\
Smooth L1 & 1 & 19.65 & 7.46 & 0.57 & 2.45 \\
Smooth L1 & 10 & 20.63 & 11.82 & 0.61 & 2.83 \\ \hline
L2 & 0 & 19.62 & 4.82 & 0.26 & 2.34\\
L2 & 1 & 21.60 & 10.76 & 0.68 & 2.57\\
L2 & 10 & 21.62 & 2.89 & 0.57 & 2.50\\
\end{tabular}
}
\caption{Ablations on losses used for positive samples and regularization factor on negative samples of RAFs. $\mathrm{AP}_{50}$ and SGDet results are reported using HRNetW32-1S.}
\label{tab:loss}
\end{table}
We refer the loss applied at locations having GT RAFs defined as positive loss $\mathcal{L}_{raf}^{+}$, and we test different regression losses. As for locations where $\mathbf{F}_{p,x,y} = \mathbf{0}$, we apply so called negative loss $\mathcal{L}_{raf}^{-}$ using L1 for regression. $\mathcal{L}_{raf}^{-}$ will be multiplied by a factor $\beta$ for adjusting the penalty. Spatially, $\mathcal{L}_{raf}$ can be re-written as $\mathcal{L}_{raf} = \mathcal{L}_{raf}^{+} + \beta \mathcal{L}_{raf}^{-} $. Table~\ref{tab:loss} shows the effects of different losses and penalty factor on the performance. We observe better performance using L1 and $\beta$ = 10. However, when only supervise on $\mathcal{L}_{raf}^{+}$ loss, the model has comparable mean recall results and it can detect more semantic and rare relationships. On the other hand, adding $\mathcal{L}_{raf}^{-}$ loss will push the model more biased to dominating predicates like \texttt{ON} and \texttt{HAS}.

\subsubsection{Architecture Choices}
\begin{table}[h]
\centering
\resizebox{0.8\columnwidth}{!}{%
\begin{tabular}{l|l|c| Z{3em} Z{3em} Z{3em}}
 Neck & Norm & $\mathrm{AP}_{50}$ & R@50 & zR@50 & mR@50 \\
\hline 
FPN & GN & 22.75 & 11.29 & 0.71 & 2.95 \\
FPN & MS-BN & 22.10 & 13.23 & 0.75 & 2.67 \\
FPN{\scriptsize$\times 2$} & GN & 22.74 & 11.96 & 0.78 & 3.00\\
FPN{\scriptsize$\times 2$} & MS-BN & 22.60 & 12.01 & 0.80 & 2.88 \\
\end{tabular}
}
\caption{Comparisons of FPN \vs FPN{\scriptsize$\times 2$}, and Multi-scale batch normalization \vs group normalization. $\mathrm{AP}_{50}$ and SGDet results are reported using ResNet50-4S.}
\label{tab:fpn}
\end{table}

By comparing our results between single-scale and multi-scale models, we see substantial performance gain on both object detection and scene graph generation from Table~\ref{tab:recall} \ref{tab:mrecall} \ref{tab:zsr}. We also observe HRNet has better results over ResNet due to its multi-scale feature fusions. For investigating the entanglement of object features and contextual features producing RAFs, we compare the results of FPN and FPN{\scriptsize$\times 2$} using ResNet-50 as backbone shown in Table~\ref{tab:fpn}. As observed in~\cite{tian2019fcos}, the regression range differs across different levels. Therefore, to improve the performance of shared fully-convolutional heads, we replace each batch normalization (BN)~\cite{ioffe2015batch} layer in the head with a set of BN layers, each of which is only applied for the corresponding scale. We name this modified BN as multi-scale batch normalization (MS-BN). We also experiment on group normalization (GN)~\cite{wu2018group} which stabilizes the training as well. We show the comparisons of MS-BN and GN (Section~\ref{sec:ms}) in the same table. We observe better mR by using FPN{\scriptsize$\times 2$} and better recall and zsR by using MS-BN. We expect more performance improvement if using larger batch size with MS-BN.

\newcolumntype{R}[1]{w{r}{#1}}
\begin{table}[t]
\centering
\resizebox{0.9\linewidth}{!}{%
\begin{tabular}{l|R{2em}|l|c}
\hline 

Method & \multicolumn{1}{c|}{\#Params\,(M)} & \multicolumn{1}{c|}{Input Size} & s / image \\\hline 

Pixels2Graphs \cite{zhang2017visual} & 94.8 & $512 \times 512$ & 3.55 \\
VCTree-TDE \cite{tang2020unbiased} & 360.8 & $600 \times 1000$		 & 1.69 \\
MOTIFS-TDE \cite{tang2020unbiased} & 369.5 & $600 \times 1000$ & 0.87 \\
KERN \cite{chen2019knowledge} & 405.2 & $592 \times 592$ & 0.79 \\
MOTIFS \cite{tang2020unbiased} & 367.2 & $600 \times 1000$ & 0.66 \\
FactorizableNet~\cite{li2018factorizable} & 40.4 & $600 \times 1000$ & 0.59 \\
VTransE-TDE \cite{tang2020unbiased} & 311.6 & $600 \times 1000$ & 0.55 \\
GB-N{\footnotesize ET}-$\beta$~\cite{zareian2020bridging} & 444.6 & $592 \times 592$ & 0.52 \\
Graph R-CNN \cite{yang2018graph} & 80.2 & $ 800 \times 1024$ & 0.19 \\ \hline
\multicolumn{4}{l}{\textbf{FCSGG} (\textbf{Ours})} \\\hline
HRNetW32-1S & 47.3 & $512 \times 512$ & 0.07 \\
HRNetW48-1S & 86.1 & $512 \times 512$ & 0.08 \\
%ResNet50-4S-FPN & & - & - \\
ResNet50-4S-FPN{\scriptsize$\times 2$} & 36.0 & $512 \times 512$ & 0.04 \\
%HRNetW32-4S-FPN{\scriptsize$\times 2$} & - & - & - \\
HRNetW48-5S-FPN{\scriptsize$\times 2$} & 87.1 & $640 \times 1024$ & 0.12 \\
HRNetW48-5S-FPN{\scriptsize$\times 2$}-\textit{f} & 87.1 & $640 \times 1024$ & 0.12 \\
\hline
\end{tabular}
}
\caption{Model size and speed comparisons for SGDet.}
\label{tab:speed}
\end{table}
\subsubsection{Model Size and Speed}
We also conduct experiments on the model size and inference speed. Few work benchmarked on efficiency of scene graph generation previously~\cite{li2018factorizable,zareian2020bridging}. Though scene graphs are powerful, it is almost not possible to perform SGG and down-stream tasks in real-time due to significantly increased model complexity. FCSGG alleviates the computational complexity effectively. Our experiments are performed on a same NVIDIA GeForce GTX 1080 Ti GPU with inference batch size of 1. For comparisons, we include several previous work by running corresponding open-source codes under the same settings. The results are shown in Table~\ref{tab:speed}. Both the number of parameters and inference time are considerably lower for FCSGG models. It is worth noting that the computation overhead is from the backbone network. The path integral (Algorithm.~\ref{alg:line_integral}) is performed pair-wisely for all 100 kept objects across five scales, which results in ${100 \choose{2}} \times 2 \times 5 = 49500$ maximum number of candidate relationships for an image. The inference time for path integral is almost invariant over the number of instances as analyzed by Cao~\etal~\cite{cao2019openpose}. We believe that object relationships exist universally, especially geometric ones. By grounding the full graph in RAFs as intermediate features, richer semantics can be retained for down-stream tasks. More importantly, convolution is hardware-friendly, and the model size is kept small for deployment on edge devices. We anticipate that real-time mobile SGG can be performed in the near future.

%------------------------------------------------------------------------

\section{Conclusions} \label{sec:conclusions}

Scene graph generation is a critical pillar for building machines to visually understand scenes and perform high-level vision and language tasks. In this paper, we introduce a fully convolutional scene graph generation framework that is simple yet effective with fast inference speed. The proposed relation affinity fields serve as a novel representation for visual relationship and produce strong generalizability for unseen relationships. By only using visual features, our exploratory method achieves competitive results over object detection and SGG metrics on the VG dataset. We expect that FCSGG can serve as a general and strong baseline for SGG task, as well as a vital building block extending to down-stream tasks.

\section*{Acknowledgement}

% \noindent\textbf{Acknowledgement}. 
This work was partially supported by Bourns Endowment funds at the University of California, Riverside.

{\small
\bibliographystyle{ieee_fullname}
\bibliography{egbib}

\begin{thebibliography}{10}\itemsep=-1pt

\bibitem{baker2011database}
Simon Baker, Daniel Scharstein, JP Lewis, Stefan Roth, Michael~J Black, and
  Richard Szeliski.
\newblock A database and evaluation methodology for optical flow.
\newblock {\em International journal of computer vision}, 92(1):1--31, 2011.

\bibitem{benjamin}
W. Benjamin and E. Leslie.
\newblock {\em On Photography}.
\newblock Reaktion Books, 2015.

\bibitem{berger}
J. Berger.
\newblock {\em Ways of Seeing}.
\newblock Penguin Modern Classics. Penguin Books Limited, 2008.

\bibitem{cao2019openpose}
Zhe Cao, Gines~Hidalgo Martinez, Tomas Simon, Shih-En Wei, and Yaser~A Sheikh.
\newblock Openpose: Realtime multi-person 2d pose estimation using part
  affinity fields.
\newblock {\em IEEE Transactions on Pattern Analysis and Machine Intelligence},
  2019.

\bibitem{chen2019counterfactual}
Long Chen, Hanwang Zhang, Jun Xiao, Xiangnan He, Shiliang Pu, and Shih-Fu
  Chang.
\newblock Counterfactual critic multi-agent training for scene graph
  generation.
\newblock In {\em Proceedings of the IEEE International Conference on Computer
  Vision}, pages 4613--4623, 2019.

\bibitem{chen2019knowledge}
Tianshui Chen, Weihao Yu, Riquan Chen, and Liang Lin.
\newblock Knowledge-embedded routing network for scene graph generation.
\newblock In {\em Proceedings of the IEEE Conference on Computer Vision and
  Pattern Recognition}, pages 6163--6171, 2019.

\bibitem{chen2019scene}
Vincent~S Chen, Paroma Varma, Ranjay Krishna, Michael Bernstein, Christopher
  Re, and Li Fei-Fei.
\newblock Scene graph prediction with limited labels.
\newblock In {\em Proceedings of the IEEE International Conference on Computer
  Vision}, pages 2580--2590, 2019.

\bibitem{cho2014properties}
Kyunghyun Cho, Bart van Merri{\"e}nboer, Dzmitry Bahdanau, and Yoshua Bengio.
\newblock On the properties of neural machine translation: Encoder--decoder
  approaches.
\newblock {\em Syntax, Semantics and Structure in Statistical Translation},
  page 103, 2014.

\bibitem{chomsky_2006}
Noam Chomsky.
\newblock {\em Language and Mind}.
\newblock Cambridge University Press, 3 edition, 2006.

\bibitem{dai2017detecting}
Bo Dai, Yuqi Zhang, and Dahua Lin.
\newblock Detecting visual relationships with deep relational networks.
\newblock In {\em Proceedings of the IEEE conference on computer vision and
  Pattern recognition}, pages 3076--3086, 2017.

\bibitem{dornadula2019visual}
Apoorva Dornadula, Austin Narcomey, Ranjay Krishna, Michael Bernstein, and
  Fei-Fei Li.
\newblock Visual relationships as functions: Enabling few-shot scene graph
  prediction.
\newblock In {\em Proceedings of the IEEE International Conference on Computer
  Vision Workshops}, pages 0--0, 2019.

\bibitem{frome2013devise}
Andrea Frome, Greg~S Corrado, Jon Shlens, Samy Bengio, Jeff Dean, Marc'Aurelio
  Ranzato, and Tomas Mikolov.
\newblock Devise: A deep visual-semantic embedding model.
\newblock In {\em Advances in neural information processing systems}, pages
  2121--2129, 2013.

\bibitem{girshick2015fast}
Ross Girshick.
\newblock Fast r-cnn.
\newblock In {\em Proceedings of the IEEE international conference on computer
  vision}, pages 1440--1448, 2015.

\bibitem{gkanatsios2019attention}
Nikolaos Gkanatsios, Vassilis Pitsikalis, Petros Koutras, and Petros Maragos.
\newblock Attention-translation-relation network for scalable scene graph
  generation.
\newblock In {\em Proceedings of the IEEE International Conference on Computer
  Vision Workshops}, pages 0--0, 2019.

\bibitem{Gkanatsios_2019_ICCV}
Nikolaos Gkanatsios, Vassilis Pitsikalis, Petros Koutras, and Petros Maragos.
\newblock Attention-translation-relation network for scalable scene graph
  generation.
\newblock In {\em Proceedings of the IEEE/CVF International Conference on
  Computer Vision (ICCV) Workshops}, Oct 2019.

\bibitem{gkanatsiossaturation}
Nikolaos Gkanatsios, Vassilis Pitsikalis, and Petros Maragos.
\newblock From saturation to zero-shot visual relationship detection using
  local context.
\newblock In {\em Proceedings of the British Machine Vision Conference}, 2020.

\bibitem{gu2019scene}
Jiuxiang Gu, Handong Zhao, Zhe Lin, Sheng Li, Jianfei Cai, and Mingyang Ling.
\newblock Scene graph generation with external knowledge and image
  reconstruction.
\newblock In {\em Proceedings of the IEEE Conference on Computer Vision and
  Pattern Recognition}, pages 1969--1978, 2019.

\bibitem{h2020iclr}
David Harwath, Wei-Ning Hsu, and James Glass.
\newblock Learning hierarchical discrete linguistic units from visually-
  grounded speech.
\newblock In {\em International Conference on Learning Representations}, 2020.

\bibitem{he2017mask}
Kaiming He, Georgia Gkioxari, Piotr Doll{\'a}r, and Ross Girshick.
\newblock Mask r-cnn.
\newblock In {\em Proceedings of the IEEE international conference on computer
  vision}, pages 2961--2969, 2017.

\bibitem{he2016deep}
Kaiming He, Xiangyu Zhang, Shaoqing Ren, and Jian Sun.
\newblock Deep residual learning for image recognition.
\newblock In {\em Proceedings of the IEEE conference on computer vision and
  pattern recognition}, pages 770--778, 2016.

\bibitem{hochreiter1997long}
Sepp Hochreiter and J{\"u}rgen Schmidhuber.
\newblock Long short-term memory.
\newblock {\em Neural computation}, 9(8):1735--1780, 1997.

\bibitem{hudson2019gqa}
Drew~A Hudson and Christopher~D Manning.
\newblock Gqa: A new dataset for real-world visual reasoning and compositional
  question answering.
\newblock In {\em Proceedings of the IEEE Conference on Computer Vision and
  Pattern Recognition}, pages 6700--6709, 2019.

\bibitem{hung2020contextual}
Zih-Siou Hung, Arun Mallya, and Svetlana Lazebnik.
\newblock Contextual translation embedding for visual relationship detection
  and scene graph generation.
\newblock {\em IEEE Transactions on Pattern Analysis and Machine Intelligence},
  2020.

\bibitem{ioffe2015batch}
Sergey Ioffe and Christian Szegedy.
\newblock Batch normalization: Accelerating deep network training by reducing
  internal covariate shift.
\newblock In {\em International Conference on Machine Learning}, pages
  448--456, 2015.

\bibitem{johnson2018image}
Justin Johnson, Agrim Gupta, and Li Fei-Fei.
\newblock Image generation from scene graphs.
\newblock In {\em Proceedings of the IEEE conference on computer vision and
  pattern recognition}, pages 1219--1228, 2018.

\bibitem{johnson2015image}
Justin Johnson, Ranjay Krishna, Michael Stark, Li-Jia Li, David Shamma, Michael
  Bernstein, and Li Fei-Fei.
\newblock Image retrieval using scene graphs.
\newblock In {\em Proceedings of the IEEE conference on computer vision and
  pattern recognition}, pages 3668--3678, 2015.

\bibitem{karpathy2015deep}
Andrej Karpathy and Li Fei-Fei.
\newblock Deep visual-semantic alignments for generating image descriptions.
\newblock In {\em Proceedings of the IEEE conference on computer vision and
  pattern recognition}, pages 3128--3137, 2015.

\bibitem{knyazev2020graph}
Boris Knyazev, Harm de Vries, C{\u{a}}t{\u{a}}lina Cangea, Graham~W Taylor,
  Aaron Courville, and Eugene Belilovsky.
\newblock Graph density-aware losses for novel compositions in scene graph
  generation.
\newblock {\em arXiv preprint arXiv:2005.08230}, 2020.

\bibitem{krishna2017visual}
Ranjay Krishna, Yuke Zhu, Oliver Groth, Justin Johnson, Kenji Hata, Joshua
  Kravitz, Stephanie Chen, Yannis Kalantidis, Li-Jia Li, David~A Shamma, et~al.
\newblock Visual genome: Connecting language and vision using crowdsourced
  dense image annotations.
\newblock {\em International journal of computer vision}, 123(1):32--73, 2017.

\bibitem{law2018cornernet}
Hei Law and Jia Deng.
\newblock Cornernet: Detecting objects as paired keypoints.
\newblock In {\em Proceedings of the European Conference on Computer Vision
  (ECCV)}, pages 734--750, 2018.

\bibitem{poeticimage}
C.D. Lewis.
\newblock {\em The Poetic Image (Clark Lectures)}.
\newblock Cambridge, 1946.

\bibitem{li2017vip}
Yikang Li, Wanli Ouyang, Xiaogang Wang, and Xiao'ou Tang.
\newblock Vip-cnn: Visual phrase guided convolutional neural network.
\newblock In {\em Proceedings of the IEEE Conference on Computer Vision and
  Pattern Recognition}, pages 1347--1356, 2017.

\bibitem{li2018factorizable}
Yikang Li, Wanli Ouyang, Bolei Zhou, Jianping Shi, Chao Zhang, and Xiaogang
  Wang.
\newblock Factorizable net: an efficient subgraph-based framework for scene
  graph generation.
\newblock In {\em Proceedings of the European Conference on Computer Vision
  (ECCV)}, pages 335--351, 2018.

\bibitem{li2017scene}
Yikang Li, Wanli Ouyang, Bolei Zhou, Kun Wang, and Xiaogang Wang.
\newblock Scene graph generation from objects, phrases and region captions.
\newblock In {\em Proceedings of the IEEE International Conference on Computer
  Vision}, pages 1261--1270, 2017.

\bibitem{lin2017feature}
Tsung-Yi Lin, Piotr Doll{\'a}r, Ross Girshick, Kaiming He, Bharath Hariharan,
  and Serge Belongie.
\newblock Feature pyramid networks for object detection.
\newblock In {\em Proceedings of the IEEE conference on computer vision and
  pattern recognition}, pages 2117--2125, 2017.

\bibitem{lin2017focal}
Tsung-Yi Lin, Priya Goyal, Ross Girshick, Kaiming He, and Piotr Doll{\'a}r.
\newblock Focal loss for dense object detection.
\newblock In {\em Proceedings of the IEEE international conference on computer
  vision}, pages 2980--2988, 2017.

\bibitem{lin2014microsoft}
Tsung-Yi Lin, Michael Maire, Serge Belongie, James Hays, Pietro Perona, Deva
  Ramanan, Piotr Doll{\'a}r, and C~Lawrence Zitnick.
\newblock Microsoft coco: Common objects in context.
\newblock In {\em European conference on computer vision}, pages 740--755.
  Springer, 2014.

\bibitem{lin2020gps}
Xin Lin, Changxing Ding, Jinquan Zeng, and Dacheng Tao.
\newblock Gps-net: Graph property sensing network for scene graph generation.
\newblock In {\em Proceedings of the IEEE/CVF Conference on Computer Vision and
  Pattern Recognition}, pages 3746--3753, 2020.

\bibitem{lu2016visual}
Cewu Lu, Ranjay Krishna, Michael Bernstein, and Li Fei-Fei.
\newblock Visual relationship detection with language priors.
\newblock In {\em European conference on computer vision}, pages 852--869.
  Springer, 2016.

\bibitem{mortazavi2020}
Masood Mortazavi.
\newblock Speech-image semantic alignment does not depend on any prior
  classification tasks.
\newblock In {\em Proceedings of InterSpeech}, 2020.

\bibitem{newell2017pixels}
Alejandro Newell and Jia Deng.
\newblock Pixels to graphs by associative embedding.
\newblock In {\em Advances in neural information processing systems}, pages
  2171--2180, 2017.

\bibitem{plummer2017phrase}
Bryan~A Plummer, Arun Mallya, Christopher~M Cervantes, Julia Hockenmaier, and
  Svetlana Lazebnik.
\newblock Phrase localization and visual relationship detection with
  comprehensive image-language cues.
\newblock In {\em Proceedings of the IEEE International Conference on Computer
  Vision}, pages 1928--1937, 2017.

\bibitem{ren2015faster}
Shaoqing Ren, Kaiming He, Ross Girshick, and Jian Sun.
\newblock Faster r-cnn: Towards real-time object detection with region proposal
  networks.
\newblock In {\em Advances in neural information processing systems}, pages
  91--99, 2015.

\bibitem{rumelhart1986learning}
David~E Rumelhart, Geoffrey~E Hinton, and Ronald~J Williams.
\newblock Learning representations by back-propagating errors.
\newblock {\em nature}, 323(6088):533--536, 1986.

\bibitem{rylemind}
Gilbert Ryle.
\newblock {\em The Concept of Mind}.
\newblock U of Chicago Press, 1949.

\bibitem{scarselli2008graph}
Franco Scarselli, Marco Gori, Ah~Chung Tsoi, Markus Hagenbuchner, and Gabriele
  Monfardini.
\newblock The graph neural network model.
\newblock {\em IEEE Transactions on Neural Networks}, 20(1):61--80, 2008.

\bibitem{sontag}
S. Sontag.
\newblock {\em On Photography}.
\newblock Kushiel's Legacy. Picador, 2001.

\bibitem{tai2015improved}
Kai~Sheng Tai, Richard Socher, and Christopher~D Manning.
\newblock Improved semantic representations from tree-structured long
  short-term memory networks.
\newblock In {\em Proceedings of the 53rd Annual Meeting of the Association for
  Computational Linguistics and the 7th International Joint Conference on
  Natural Language Processing (Volume 1: Long Papers)}, pages 1556--1566, 2015.

\bibitem{tan2020efficientdet}
Mingxing Tan, Ruoming Pang, and Quoc~V Le.
\newblock Efficientdet: Scalable and efficient object detection.
\newblock In {\em Proceedings of the IEEE/CVF conference on computer vision and
  pattern recognition}, pages 10781--10790, 2020.

\bibitem{tang2020unbiased}
Kaihua Tang, Yulei Niu, Jianqiang Huang, Jiaxin Shi, and Hanwang Zhang.
\newblock Unbiased scene graph generation from biased training.
\newblock In {\em Proceedings of the IEEE/CVF Conference on Computer Vision and
  Pattern Recognition}, pages 3716--3725, 2020.

\bibitem{tang2019learning}
Kaihua Tang, Hanwang Zhang, Baoyuan Wu, Wenhan Luo, and Wei Liu.
\newblock Learning to compose dynamic tree structures for visual contexts.
\newblock In {\em Proceedings of the IEEE Conference on Computer Vision and
  Pattern Recognition}, pages 6619--6628, 2019.

\bibitem{teney2017graph}
Damien Teney, Lingqiao Liu, and Anton van Den~Hengel.
\newblock Graph-structured representations for visual question answering.
\newblock In {\em Proceedings of the IEEE conference on computer vision and
  pattern recognition}, pages 1--9, 2017.

\bibitem{tian2019fcos}
Zhi Tian, Chunhua Shen, Hao Chen, and Tong He.
\newblock Fcos: Fully convolutional one-stage object detection.
\newblock In {\em Proceedings of the IEEE international conference on computer
  vision}, pages 9627--9636, 2019.

\bibitem{wang2020deep}
Jingdong Wang, Ke Sun, Tianheng Cheng, Borui Jiang, Chaorui Deng, Yang Zhao,
  Dong Liu, Yadong Mu, Mingkui Tan, Xinggang Wang, et~al.
\newblock Deep high-resolution representation learning for visual recognition.
\newblock {\em IEEE transactions on pattern analysis and machine intelligence},
  2020.

\bibitem{wang2019exploring}
Wenbin Wang, Ruiping Wang, Shiguang Shan, and Xilin Chen.
\newblock Exploring context and visual pattern of relationship for scene graph
  generation.
\newblock In {\em Proceedings of the IEEE Conference on Computer Vision and
  Pattern Recognition}, pages 8188--8197, 2019.

\bibitem{woo2018linknet}
Sanghyun Woo, Dahun Kim, Donghyeon Cho, and In~So Kweon.
\newblock Linknet: Relational embedding for scene graph.
\newblock In {\em Advances in Neural Information Processing Systems}, pages
  560--570, 2018.

\bibitem{wu2019unified}
Hao Wu, Jiayuan Mao, Yufeng Zhang, Yuning Jiang, Lei Li, Weiwei Sun, and
  Wei-Ying Ma.
\newblock Unified visual-semantic embeddings: Bridging vision and language with
  structured meaning representations.
\newblock In {\em Proceedings of the IEEE Conference on Computer Vision and
  Pattern Recognition}, pages 6609--6618, 2019.

\bibitem{wu2018group}
Yuxin Wu and Kaiming He.
\newblock Group normalization.
\newblock In {\em Proceedings of the European conference on computer vision
  (ECCV)}, pages 3--19, 2018.

\bibitem{wu2019detectron2}
Yuxin Wu, Alexander Kirillov, Francisco Massa, Wan-Yen Lo, and Ross Girshick.
\newblock Detectron2.
\newblock \url{https://github.com/facebookresearch/detectron2}, 2019.

\bibitem{xu2017scene}
Danfei Xu, Yuke Zhu, Christopher~B Choy, and Li Fei-Fei.
\newblock Scene graph generation by iterative message passing.
\newblock In {\em Proceedings of the IEEE conference on computer vision and
  pattern recognition}, pages 5410--5419, 2017.

\bibitem{xu2018attngan}
Tao Xu, Pengchuan Zhang, Qiuyuan Huang, Han Zhang, Zhe Gan, Xiaolei Huang, and
  Xiaodong He.
\newblock Attngan: Fine-grained text to image generation with attentional
  generative adversarial networks.
\newblock In {\em Proceedings of the IEEE conference on computer vision and
  pattern recognition}, pages 1316--1324, 2018.

\bibitem{yang2018graph}
Jianwei Yang, Jiasen Lu, Stefan Lee, Dhruv Batra, and Devi Parikh.
\newblock Graph r-cnn for scene graph generation.
\newblock In {\em Proceedings of the European conference on computer vision
  (ECCV)}, pages 670--685, 2018.

\bibitem{yang2019auto}
Xu Yang, Kaihua Tang, Hanwang Zhang, and Jianfei Cai.
\newblock Auto-encoding scene graphs for image captioning.
\newblock In {\em Proceedings of the IEEE Conference on Computer Vision and
  Pattern Recognition}, pages 10685--10694, 2019.

\bibitem{yang2018shuffle}
Xu Yang, Hanwang Zhang, and Jianfei Cai.
\newblock Shuffle-then-assemble: Learning object-agnostic visual relationship
  features.
\newblock In {\em Proceedings of the European conference on computer vision
  (ECCV)}, pages 36--52, 2018.

\bibitem{yang2019reppoints}
Ze Yang, Shaohui Liu, Han Hu, Liwei Wang, and Stephen Lin.
\newblock Reppoints: Point set representation for object detection.
\newblock In {\em Proceedings of the IEEE/CVF International Conference on
  Computer Vision}, pages 9657--9666, 2019.

\bibitem{yao2018exploring}
Ting Yao, Yingwei Pan, Yehao Li, and Tao Mei.
\newblock Exploring visual relationship for image captioning.
\newblock In {\em Proceedings of the European conference on computer vision
  (ECCV)}, pages 684--699, 2018.

\bibitem{yu2017visual}
Ruichi Yu, Ang Li, Vlad~I Morariu, and Larry~S Davis.
\newblock Visual relationship detection with internal and external linguistic
  knowledge distillation.
\newblock In {\em Proceedings of the IEEE international conference on computer
  vision}, pages 1974--1982, 2017.

\bibitem{zareian2020bridging}
Alireza Zareian, Svebor Karaman, and Shih-Fu Chang.
\newblock Bridging knowledge graphs to generate scene graphs.
\newblock {\em arXiv preprint arXiv:2001.02314}, 2020.

\bibitem{zareian2020learning}
Alireza Zareian, Haoxuan You, Zhecan Wang, and Shih-Fu Chang.
\newblock Learning visual commonsense for robust scene graph generation.
\newblock {\em arXiv preprint arXiv:2006.09623}, 2020.

\bibitem{zellers2018neural}
Rowan Zellers, Mark Yatskar, Sam Thomson, and Yejin Choi.
\newblock Neural motifs: Scene graph parsing with global context.
\newblock In {\em Proceedings of the IEEE Conference on Computer Vision and
  Pattern Recognition}, pages 5831--5840, 2018.

\bibitem{zhang2017visual}
Hanwang Zhang, Zawlin Kyaw, Shih-Fu Chang, and Tat-Seng Chua.
\newblock Visual translation embedding network for visual relation detection.
\newblock In {\em Proceedings of the IEEE conference on computer vision and
  pattern recognition}, pages 5532--5540, 2017.

\bibitem{zhou2019objects}
Xingyi Zhou, Dequan Wang, and Philipp Kr{\"a}henb{\"u}hl.
\newblock Objects as points.
\newblock {\em arXiv preprint arXiv:1904.07850}, 2019.

\end{thebibliography}
}

\newpage
\appendix
\section*{Appendix}
\begin{appendices}

This supplementary document is organized as follows:
\begin {enumerate*} [1) ]%
\item more training details in Section~\ref{sec:train};
\item details on network inference in Section~\ref{sec:inference}; \item detailed architectures in Section~\ref{sec:architecture};
\item qualitative studies in Section~\ref{sec:results}.
\end {enumerate*}
All notations are followed from the paper.

%%%%%%%%% BODY TEXT
\section{Training} \label{sec:train}

% \subsection{Ground-truth Definition of Object Center}
We defined the center heatmaps in Equation 1 from the paper such that the centers are converted into 2D Gaussian masks. The standard deviations $\sigma_x$ and $\sigma_y$ are computed based on the desired radii $a^{i}$ and $b^{i}$ along x-axis and y-axis, respectively:
\begin{equation} \label{eq:radius}
	(\sigma_{x}, \sigma_{y}) = \frac{1}{3} (a^{i}, b^{i}) = \frac{1}{3} \lfloor (\sqrt{2} - 1) \, \mathbf{s}^{i}/ \tau + 1 \rfloor.
\end{equation}
The values of $a^{i}$ and $b^{i}$ are determined by Eq.(\ref{eq:radius}) such that for any point within the ellipse region of radii $a^{i}$ and $b^{i}$, when using it as a center to create a bounding box of object size $\mathbf{s}^{i}$, the bounding box has at least 0.5 intersection over union (IoU) with the GT bounding box. 

Then the prediction of center heatmaps can be supervised by forms of distance losses such as Gaussian focal loss~\cite{law2018cornernet,lin2017focal,zhou2019objects} with hyper-parameters $\alpha=2$ and $\gamma = 4$ for weight balancing. Let $\hat{\mathbf{O}}$ be the predicted center heatmap, then the pixel-wise loss $\mathcal{L}_{\mathbf{O}_{c,x,y}}$ is defined as:

\begin{equation}
\resizebox{\linewidth}{!}{$
\mathcal{L}_{\mathbf{O}_{c,x,y}}=\left\{
\begin{aligned}
 & (1 - \hat{\mathbf{O}}_{c,x,y})^\alpha \log(\hat{\mathbf{O}}_{c,x,y}) \qquad \text { if } \mathbf{O}_{c,x,y}=1 \\
&  (\hat{\mathbf{O}}_{c,x,y})^\alpha ( 1 - \mathbf{O}_{c,x,y})^\gamma \log (1 - \hat{\mathbf{O}}_{c,x,y}) \, \text { o/w. }
\end{aligned}
\right.
$}
\end{equation}

\section{Inference} \label{sec:inference}

In this section, we provide more details on inference and post-processing of our proposed fully convolutional scene graph generation (FCSGG). Our model outputs four dense feature maps: center heatmaps $\hat{\mathbf{O}}$, center offsets $\hat{\mathbf{\Delta}}$ , object sizes $\hat{\mathbf{S}}$ and relation affinity fields (RAFs) $\hat{\mathbf{F}}$. To get the object centers, we follow the same step in \cite{zhou2019objects}. Specifically, a in-place sigmoid function is applied to the predicted center heatmaps $\hat{\mathbf{O}}$ such that their values are mapped into the range of $[0, 1]$. Then a $3 \times 3$ max pooling is applied to center heatmaps for filtering duplicate detections. For a point $ \mathbf{p} = (x, y)$, the value of $\hat{\mathbf{O}}_{c,x,y}$ is considered as the measurement of the center detection score for object class $c$. Then peaks in center heatmaps are extract for each object class independently. We keep the top 100 peaks by their scores and get a set of object centers $\{\mathbf{o}^i\}^{100}_{i=1}$ with object classes $\{c^i\}^{100}_{i=1}$.

To get the corresponding center offset and object size given a detected object center $\hat{\mathbf{o}}^i = (\hat{x}^i, \hat{y}^i)$, we simply gather the values from $\hat{\mathbf{\Delta}}$ and $\hat{\mathbf{S}}$ at $\hat{\mathbf{o}}^i$. We can get the
center offset $\hat{\boldsymbol{\delta}}^i = \hat{\mathbf{\Delta}}_{\hat{x}^i,\hat{y}^i} = (\hat{\delta}^i_x, \hat{\delta}^i_y)$, and the object
size $\hat{\mathbf{s}}^i = \hat{\mathbf{S}}_{\hat{x}^i,\hat{y}^i} = (\hat{w}^i_x, \hat{h}^i_y)$. Finally, the bounding box
$(\hat{x}^i_0, \hat{y}^i_0,\hat{x}^i_1,\hat{y}^i_1)$ of object $\hat{b}^i$ can be recovered by

\begin{equation} \label{eq:bbox}
\begin{array}{l}
\hat{x}_{0}^{i}=\left(\hat{x}^{i}+\hat{\delta}_{x}^{i}-\hat{w}^{i} / 2\right) \cdot \tau \\
\hat{y}_{0}^{i}=\left(\hat{y}^{i}+\hat{\delta}_{y}^{i}-\hat{h}^{i} / 2\right) \cdot \tau \\
\hat{x}_{1}^{i}=\left(\hat{x}^{i}+\hat{\delta}_{x}^{i}+\hat{w}^{i} / 2\right) \cdot \tau \\
\hat{y}_{1}^{i}=\left(\hat{y}^{i}+\hat{\delta}_{y}^{i}+\hat{h}^{i} / 2\right) \cdot \tau,
\end{array}
\end{equation}
where $\tau$ is the stride of the output features.

As for multi-scale prediction, we gather the top 100 detected objects for each scale, then perform a per-class non-maximum suppression (NMS) and keep the top 100 boxes from all the detections, \eg, if there are 5 scales, we will keep the top 100 boxes from the 500 boxes across all the five scales.

\subsection{Path Integral}

For multi-scale RAFs, we select the valid object pairs from the kept 100 detections $\{\mathbf{o}^i\}^{100}_{i=1}$ following the rule defined in Section 4.2 from the paper. For example, we define the predicted 5-scale RAFs as $\{ \hat{\mathbf{F}}_k \}_{k=3}^7$. If the distance between $\hat{\mathbf{o}}^i$ and $\hat{\mathbf{o}}^j$ is within $[0, 64]$, then the path integral from $\hat{\mathbf{o}}^i$ to $\hat{\mathbf{o}}^j$ will be only performed on $\hat{\mathbf{F}}_3$. We gather the top 100 relationships from each scale, and keep the top 100 relationships across all scales for evaluation.

Mentioned in the paper, the path integral is performed using matrix multiplication in practice. Specifically, we determine the longest integral “length” $m_{\mathrm{max}} = \lceil \mathrm{MAX}(\{ m^{i \to j} | \forall i \neq j \}) \rceil $ among the predicted object centers $\{ \hat{\mathbf{o}}^i \}$. In other words, there will be $m_{\mathrm{max}}$ sampled points along the integral path for each pair of object centers regardless of their distance.

\subsection{Performance Upper Bound}

One may concern the relation affinity field representation can actually work and reconstruct relationship successfully by path integral. We analyze the performance upper bound by using the ground-truth of objects and RAFs for evaluation on the test set. It achieves 91.13 R@20 and 86.85 mR@20, which proves that our proposed method is capable of recovering scene graphs from our definition of RAFs. It is worth noting that it is not possible to get 100\% re-call since there exist multiple edges between nodes in some ground-truth annotations.
%-------------------------------------------------------------------------

\section{Detailed Architectures} \label{sec:architecture}

\begin{table*}[t]
\centering
\resizebox{\textwidth}{!}{%
\begin{tabular}{l|lr|lr|llr}
\hline FCSGG & Backbone & \#Params & Neck & \#Params & Object detection heads & Relation detection head & \#Params \\
\hline 
HRNetW32-1S & Figure~\ref{fig:hrnet}, C=32 & 29.3M & Figure~\hyperref[fig:neck]{2c} & 0.0M & 256 - 256 - 256 - 256 & 512$\searrow$ - 512 - 512 - 512 - $\nearrow$ & 18.0M \\

HRNetW48-1S & Figure~\ref{fig:hrnet}, C=48 & 65.3M & Figure~\hyperref[fig:neck]{2c} & 0.0M & 256 - 256 - 256 - 256 & 512$\searrow$ - 512 - 512 - 512 - $\nearrow$ & 20.8M \\

ResNet50-4S-FPN{\scriptsize$\times 2$} & Figure~\ref{fig:resnet}, C=256 & 23.6M & Figure~\hyperref[fig:neck]{2a} & 11.4M & 64 - 64 - 64 - 64 & 64 - 64 - 64 - 64 & 1.1M \\

HRNetW48-5S-FPN{\scriptsize$\times 2$} & Figure~\ref{fig:hrnet}, C=48 & 65.3M & Figure~\hyperref[fig:neck]{2b} & 6.3M & 256 - 256 - 256 - 256 & 512 - 512 - 512 - 512 & 15.5M \\
\hline
\end{tabular}
}
\caption{FCSGG model zoo. The detailed architectures are described in corresponding figures and tables. The symbol C represents the number of feature channels in C2. Columns 6 and 7 show the number of channels for each convolution in heads. Notation $\searrow$ denotes that the convolution is of stride 2, and $\nearrow$ is bilinear interpolation for upsampling the features to the target stride ($\frac{1}{4}$ of the input image size for HRNetW32-1S and HRNetW48-1S).}
\label{tab:architecture}
\end{table*}
In this section, we provide more details of the proposed fully convolutional scene graph generation model. Our codebase is based on Detectron2~\cite{wu2019detectron2} and Tang~\etal~\cite{tang2020unbiased}. We list the models mentioned in the paper in Table 1 again for convenience. The number of parameters (\#Params) of each network module is also listed in Table~\ref{tab:architecture}. As shown in the table, the backbone network has the largest number of parameters while the heads are relatively small.

\subsection{Backbone}

\begin{figure}[b]
     \centering
     \begin{subfigure}[b]{\linewidth}
         \centering
         \includegraphics[width=\linewidth]{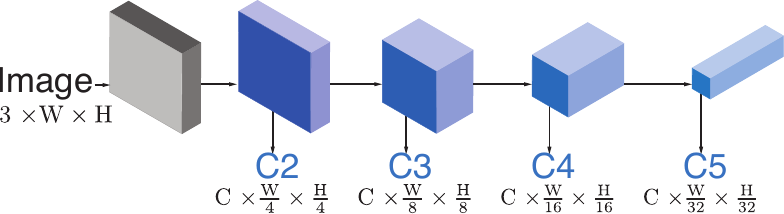}
         \caption{An example of ResNet, where $\mathrm{C} = 256$ for ResNet-50 and ResNet-101.}
         \label{fig:resnet}
     \end{subfigure}
     \hfill
     \begin{subfigure}[b]{\linewidth}
         \centering
         \includegraphics[width=\linewidth]{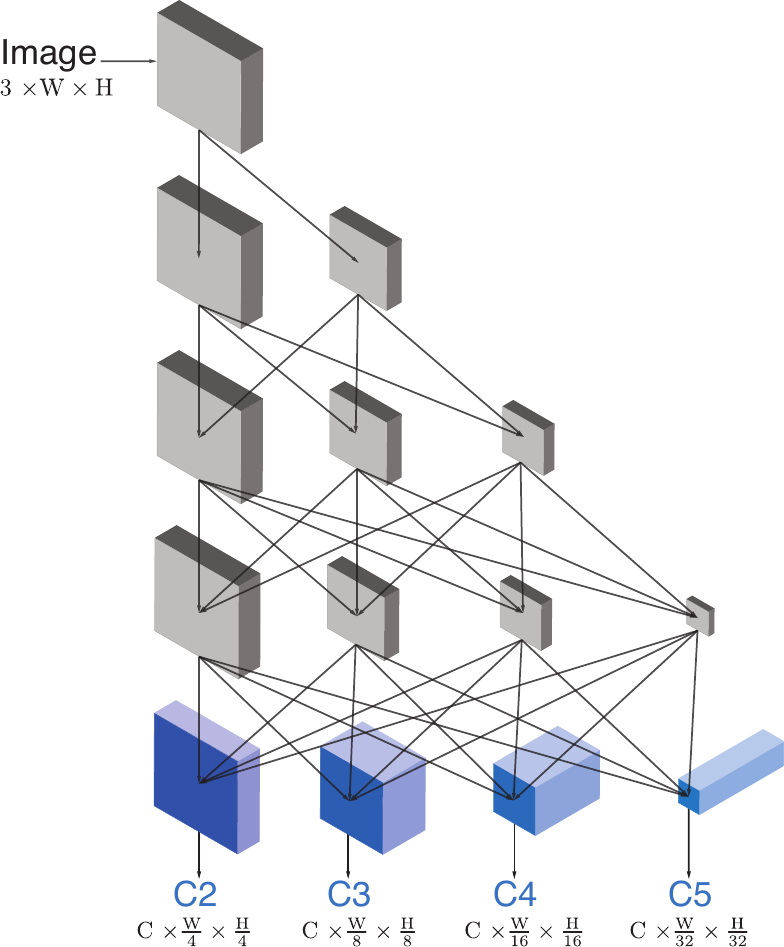}
         \caption{An example of HRNet, where $\mathrm{C} = 32$ for HRNetW32 and $\mathrm{C} = 48$ for HRNetW48.}
         \label{fig:hrnet}
     \end{subfigure}
        \caption{Conceptual backbone architectures.}
        \label{fig:backbone}
\end{figure}

The backbone network serves as a feature extraction module in most of the deep learning applications. We choose the widely used network ResNet~\cite{he2016deep}  and a recent successful alternative named HRNet~\cite{wang2020deep}. ResNet is a representative of deep networks such that the resolution of the feature maps is downsampled while the number of channels is increased, sequentially. ResNet can be divided into stages after the “stem” (first several convolutional layers of the backbone), and the output features of each stage is named as C2, C3, C4, and C5 respectively. On the other hand, HRNet maintains a higher feature resolution all the way to the network output, and constructs several branches of features with lower resolutions. Features from each branch will be fused for exchanging information repeatedly. The output features of each branch are named as C2, C3, C4, and C5 respectively for convenience. The conceptual architecture diagrams of ResNet and HRNet are shown in the figure 1. We do not change the architecture and hyper-parameters of the backbone network with respect to the original papers~\cite{he2016deep,wang2020deep}.

\subsection{Neck}

\begin{figure}[t]
    \centering
    \includegraphics[width=\linewidth]{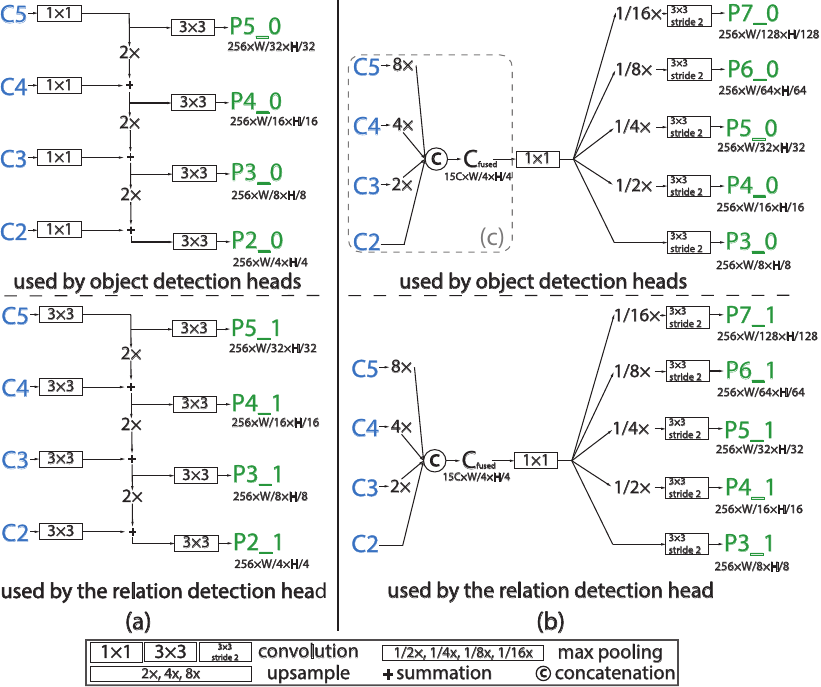}
    \caption{Conceptual neck architectures. The left sub-figure (a) represents the FPN, and the right sub-figure (b) represents HRNetV2p with five scales of features. The dashed block (c) represents the feature fusion module for the single-scale HRNet.}
    \label{fig:neck}
\end{figure}

As presented in the paper, the neck networks serves as a module for constructing multiple scales of features that can be used for later predictions. We use feature pyramid network (FPN)~\cite{lin2017feature} as the neck for ResNet, which is widely used for object detection~\cite{he2017mask,lin2017focal,tian2019fcos}. FPN allows information exchange across different scales of features after backbone feature extraction. By up-sampling higher level of features (\eg C5) then summing with lower level of features (\eg output features from C4 after a 1 convolution) consecutively, a pyramid of feature maps (with the same number of channels) is built and called \{P2, P3, P4, P5\}. For ResNet50-4S-FPN$_{\times 2}$, we use a modified version of FPN called bidirectional FPN (BiFPN~\cite{tan2020efficientdet}) which allows more connections among each scale. A detailed illustration of FPN as a neck is shown in Figure~\hyperref[fig:neck]{2a}.

As for HRNet as backbone, we follow Wang~\etal~\cite{wang2020deep} and use the HRNetV2 for single-scale prediction, and HR-NetV2p network for multi-scale feature representations. It should be addressed that even for single-scale models like HRNetW32-1S and HRNetW48-1S, a neck is applied for merging features from all branches. In this case, the neck is simply a feature fusion module without any trainable parameters. Features of C2, C3, C4, and C5 will be upsampled to the resolution of C2 via bilinear interpolation then concatenated. We name the fused features as $C_{\text{fused}}$. Different from HRNetV2~\cite{wang2020deep}, we did not use $1 \times 1$ convolution after fusion, and the resulting features will be fed into the heads. As for HRNetV2p, the first step is the same as HRNetV2, then a max pooling and a $3 \times 3$ convolution are applied on $C_{\text{fused}}$ with different strides to construct multiple scales of feature maps. The designs of necks for HRNet are shown in Figure~\hyperref[fig:neck]{2b}.

\subsection{Heads}

There are in total of four heads, and each head is responsible for the task of predicting center heatmaps, center off-sets, object size and relation affinity fields respectively. We name the first three heads as object detection heads, and the last one as the relation detection head. As stated in the paper, all heads are small network with four convolutional blocks, each of which consists of a $3 \times 3$ convolution, a normalization layer of choice such as group normalization (GN), batch normalization (BN) or multi-scale batch normalization (MS-BN), and a ReLU activation layer. Then, for each head, there is a $1 \times 1$ convolution as the output layer, and the number of channels is $\mathcal{C}$, 2, 2 and $\mathcal{P}$ respectively. The number of channels are the same among object detection heads except the output layer, while we increase the number of channels for relation detection head in some models. We list the number of channels in each block for object detection heads and relation detection head as shown in Table~\ref{tab:architecture} (the output convolution is omitted).

\section{Qualitative Results} \label{sec:results}

\begin{figure}[b]
     \centering
     \begin{subfigure}[b]{0.49\linewidth}
         \centering
         \includegraphics[width=\linewidth]{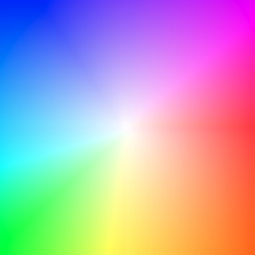}
         \caption{Field color coding.}
         \label{fig:flow_a}
     \end{subfigure}
     \hfill
     \begin{subfigure}[b]{0.49\linewidth}
         \centering
         \includegraphics[width=\linewidth]{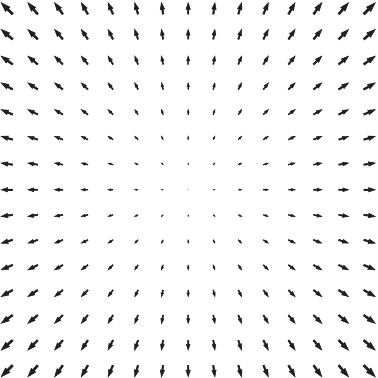}
         \caption{The corresponding vectors.}
         \label{fig:flow_b}
     \end{subfigure}
        \caption{The color coding for RAFs.}
        \label{fig:flow}
\end{figure}

We adopt the visualization method for optical flow~\cite{baker2011database} on visualizing relation affinity fields. Shown in Figure~\ref{fig:flow}, the vector orientation is represented by color hue while the vector length is encoded by color saturation. We only visualize the bounding boxes with predicted scores over 0.2, and relationships with scores over 0.1.

The visualizations of scene graph detection (SGDet) results for several test images from Visual Genome dataset~\cite{krishna2017visual} are shown in Figure~\ref{fig:vis}. From the figure, FCSGG has strong object detection performance, especially on the localization of bounding boxes. In Figure~\ref{fig:vis_a}, \texttt{pole} can be recognized even though it is not annotated in the ground-truth. In Figure~\ref{fig:vis_f}, FCSGG detects more and accurate objects compared to the ground-truth.

However, there are two challenges for training on Visual Genome dataset: object class ambiguity and predicate ambiguity. For the first challenge, as an example of Figure~\ref{fig:vis_a}, Jacket is misclassified as \texttt{coat} which is reasonable since the semantic difference between the two is subtle. Meanwhile, it also detects the person instance with even better bounding box than the ground-truth. However, it misclassified \texttt{man} as \texttt{woman}. As a result, all the relationships associated with \texttt{man} will be false detections. Similarly in Figure~\ref{fig:vis_e}, \texttt{woman} is misclassified as \texttt{lady}. We argue that these person-centric relationships take a large proportion in the Visual Genome dataset, and it is difficult to visually distinguish among person entities of similar semantics such as \texttt{woman} / \texttt{lady}, \texttt{boy} / \texttt{kid}, \texttt{man} / \texttt{men}, and \texttt{person} / \texttt{people}. Even though, FCSGG achieves superior object detection performance on Visual Genome dataset.

The other challenge is the predicate ambiguity. Even though VG-150 only keeps the top 50 frequent predicates, there are still predicates with similar semantics (\eg \texttt{OF} / \texttt{PART OF}, \texttt{WEARING} / \texttt{WEARS}, \texttt{LAYING ON} / \texttt{LYING ON}), or with vague and trivial meanings (\eg \texttt{OF}, \texttt{TO}, \texttt{NEAR}, \texttt{WITH}). FCSGG is still able to capture similar semantics with similar responses. For example, we see similar RAFs predictions between \texttt{WEARING} and \texttt{WEARS} (Figure~\ref{fig:vis_a} and \ref{fig:vis_d}). Since we do not add any predicate-specific hyper-parameters or statistic bias during training, there will always be some loss contributed by the predicate ambiguities that causes the training even harder. However, FCSGG achieves strong generalization on relationship prediction. In Figure~\ref{fig:vis_c}, it predicts \texttt{<man, RIDING ,motorcycle>}, \texttt{<person, SITTTING ON, motorcycle>}, and \texttt{motorcycle, UNDER, person>} concurrently though neither of these are annotated in the dataset.

It should be addressed that multiple predicates could be all valid for a pair of objects, \eg both \texttt{<person, ON, street>} and \texttt{<person, STANDING ON, street>} can represent the correct relationship. Our proposed RAFs are suitable for multi-class problem so that our no-graph constraint results are much improved. More importantly, \texttt{<street, UNDER, person>} is actually true even though we rarely describe this way due to language bias. Interestingly, FCSGG generalizes the relationships and learns the reciprocal correlations between predicates. From the visualizations in Figure~\ref{fig:vis_a} and \ref{fig:vis_d}, \texttt{ABOVE} and \texttt{UNDER} will have responses with similar vector magnitudes but opposite directions. We can also see the similar pattern between \texttt{OF} and \texttt{HAS}.

\begin{figure}[t!]
    \centering
    \includegraphics[width=.9\linewidth]{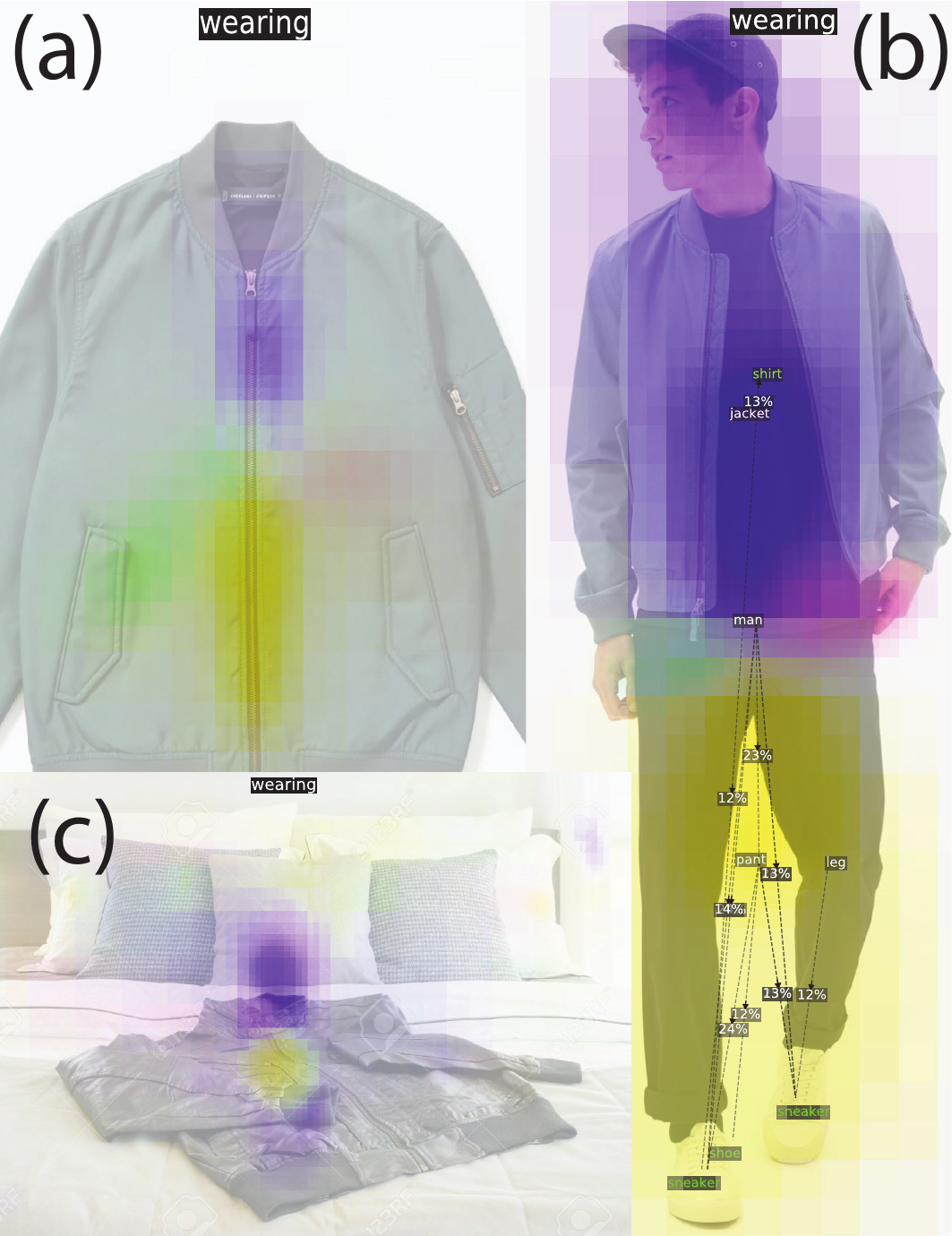}
    \caption{The RAF visualizations of \texttt{WEARING} for wild images. Among the images, (b) has the largest RAF vector norm before normalization.}
    \label{fig:jacket}
\end{figure}

\begin{figure*}[b]
     \centering
     \begin{subfigure}[b]{\linewidth}
         \centering
         \includegraphics[width=\linewidth]{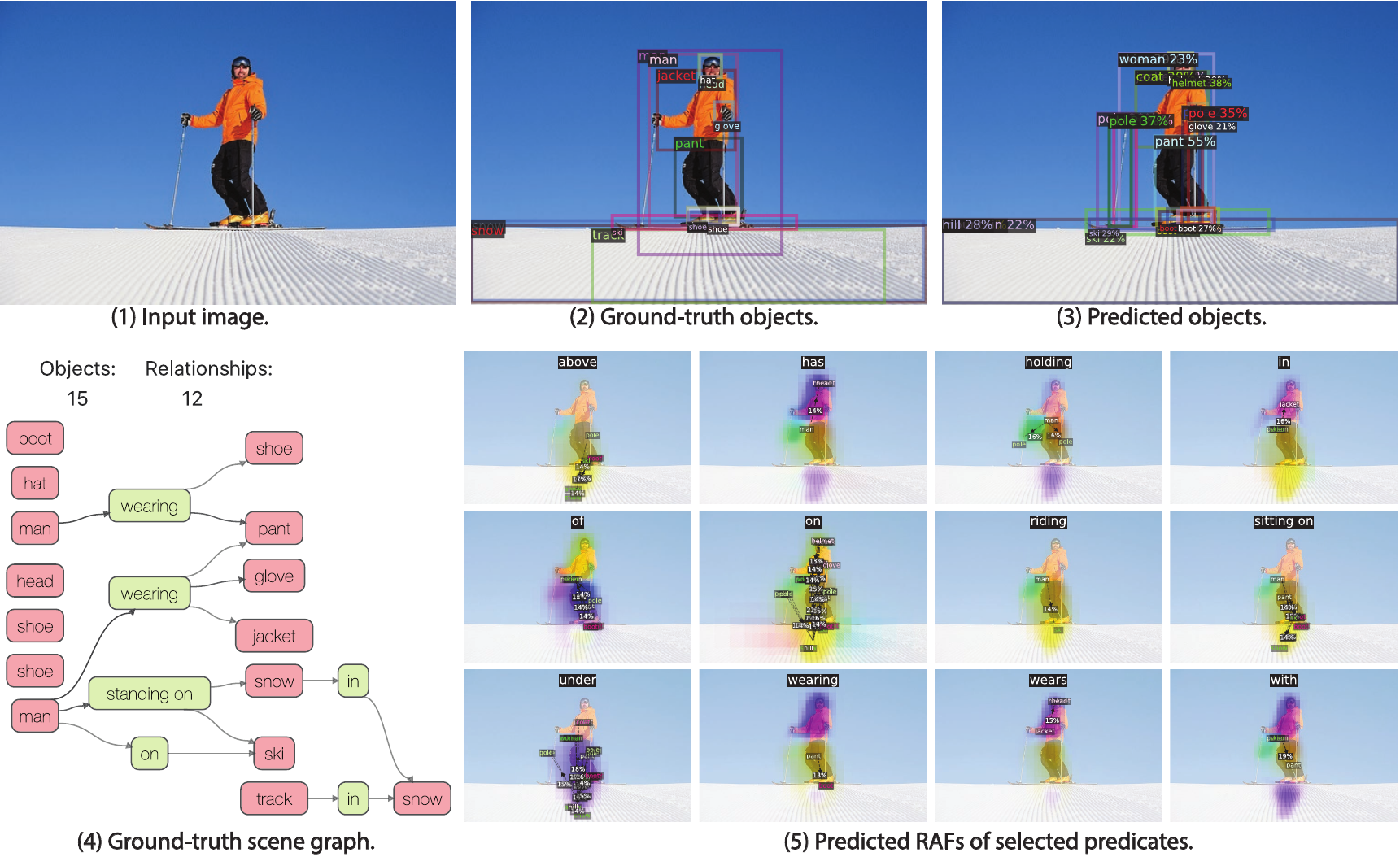}
         \caption{Visualizations of test image \texttt{2343157.jpg}.}
         \label{fig:vis_a}
     \end{subfigure}
     \hfill
     \begin{subfigure}[b]{\linewidth}
         \centering
         \includegraphics[width=\linewidth]{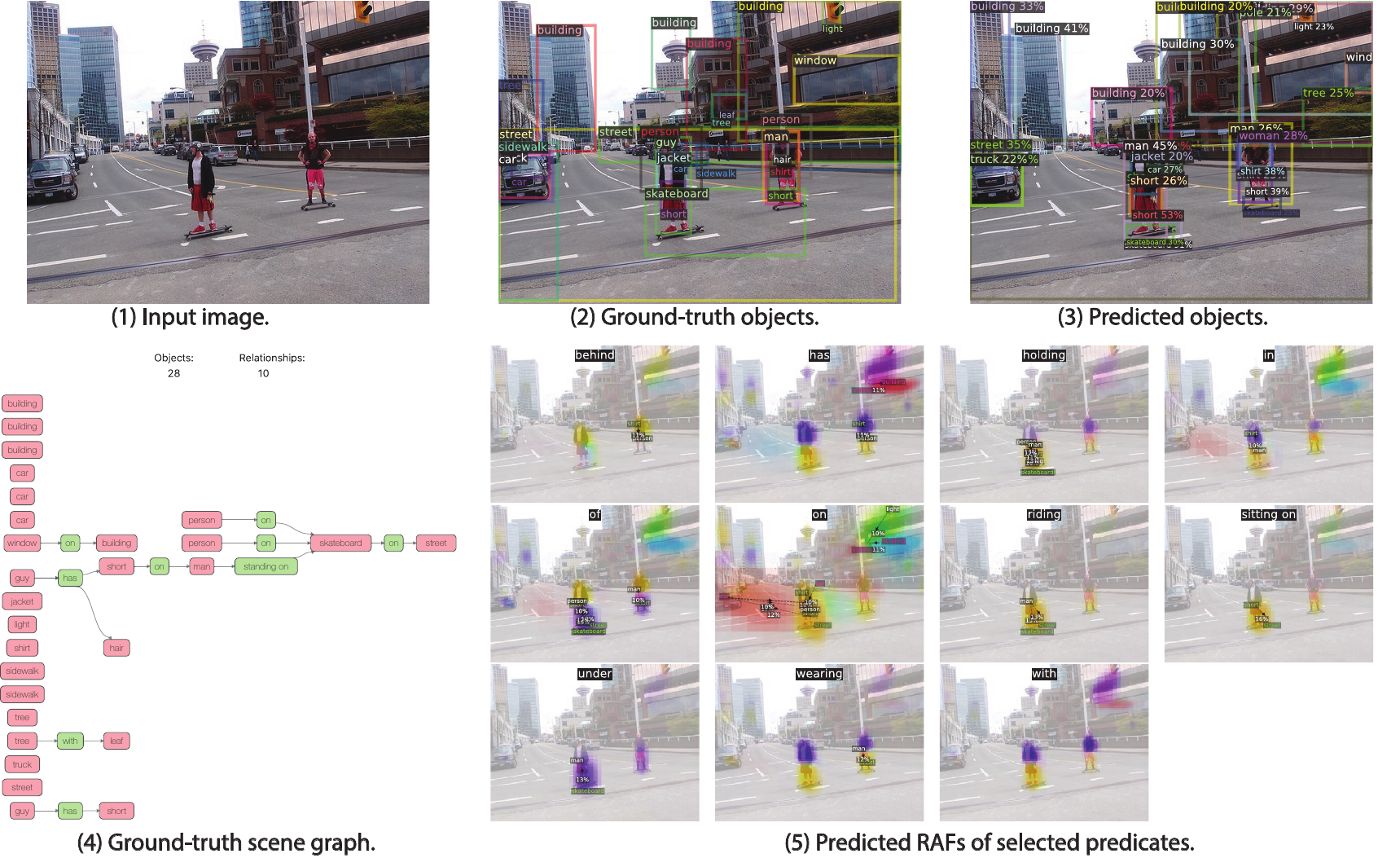}
         \caption{Visualizations of test image \texttt{2343530.jpg}.}
         \label{fig:vis_b}
     \end{subfigure}
\end{figure*}
\begin{figure*}\ContinuedFloat
     \begin{subfigure}[b]{\linewidth}
         \centering
         \includegraphics[width=\linewidth]{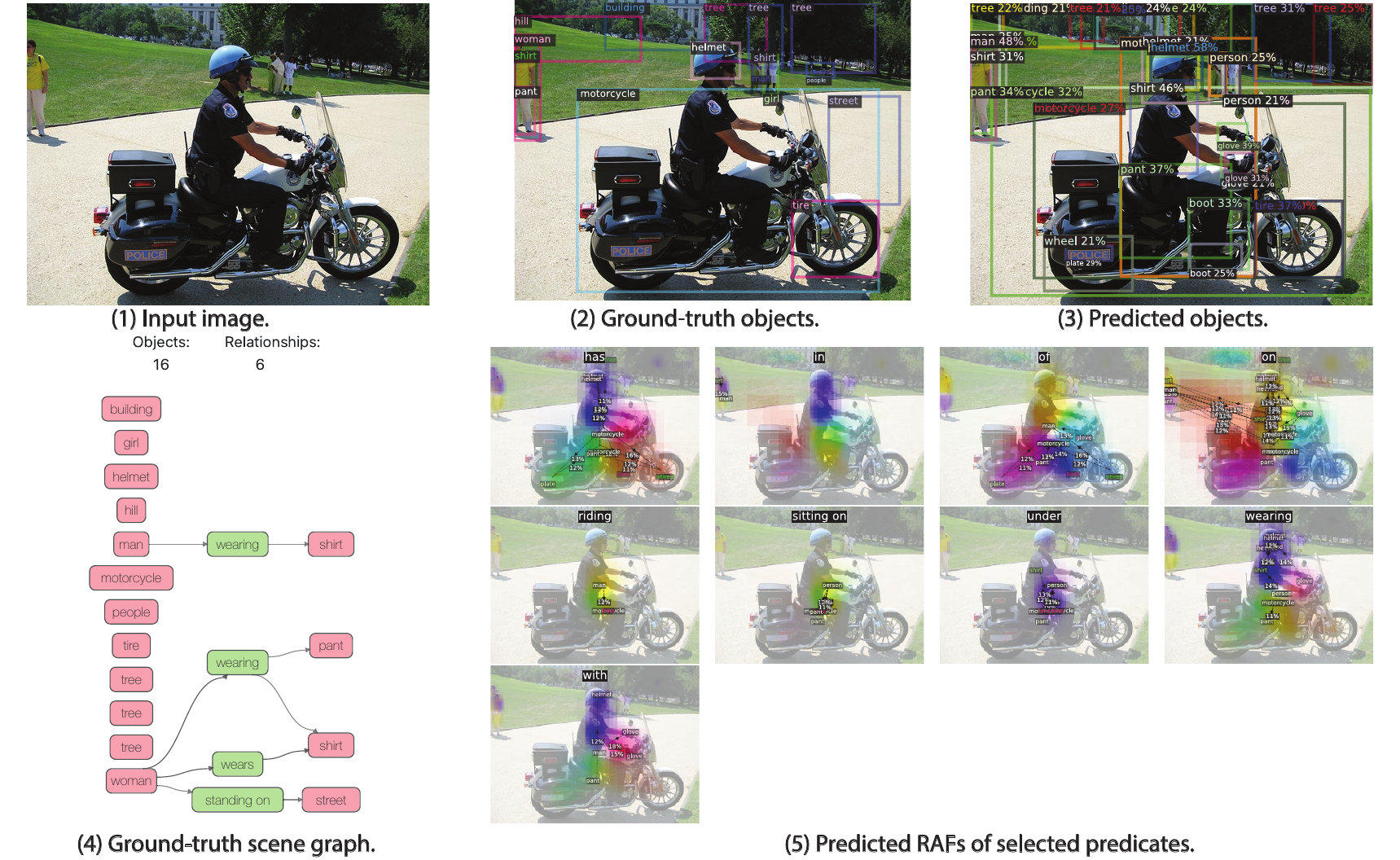}
         \caption{Visualizations of test image \texttt{2343342.jpg}.}
         \label{fig:vis_c}
     \end{subfigure}
     \hfill
     \begin{subfigure}[b]{\linewidth}
         \centering
         \includegraphics[width=\linewidth]{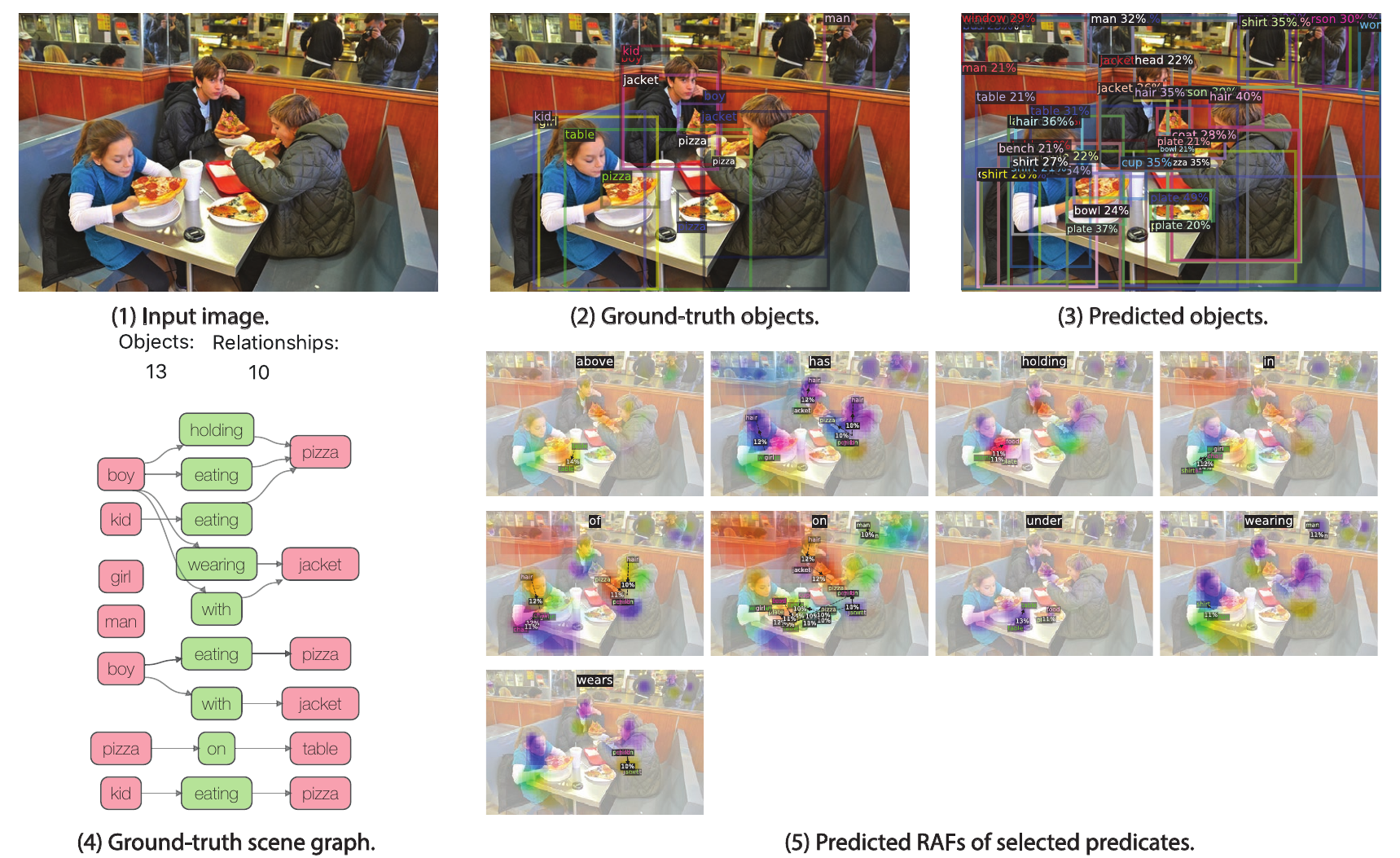}
         \caption{Visualizations of test image \texttt{2342735.jpg}.}
         \label{fig:vis_d}
     \end{subfigure}
\end{figure*}
\begin{figure*}\ContinuedFloat
     \begin{subfigure}[b]{\linewidth}
         \centering
         \includegraphics[width=\linewidth]{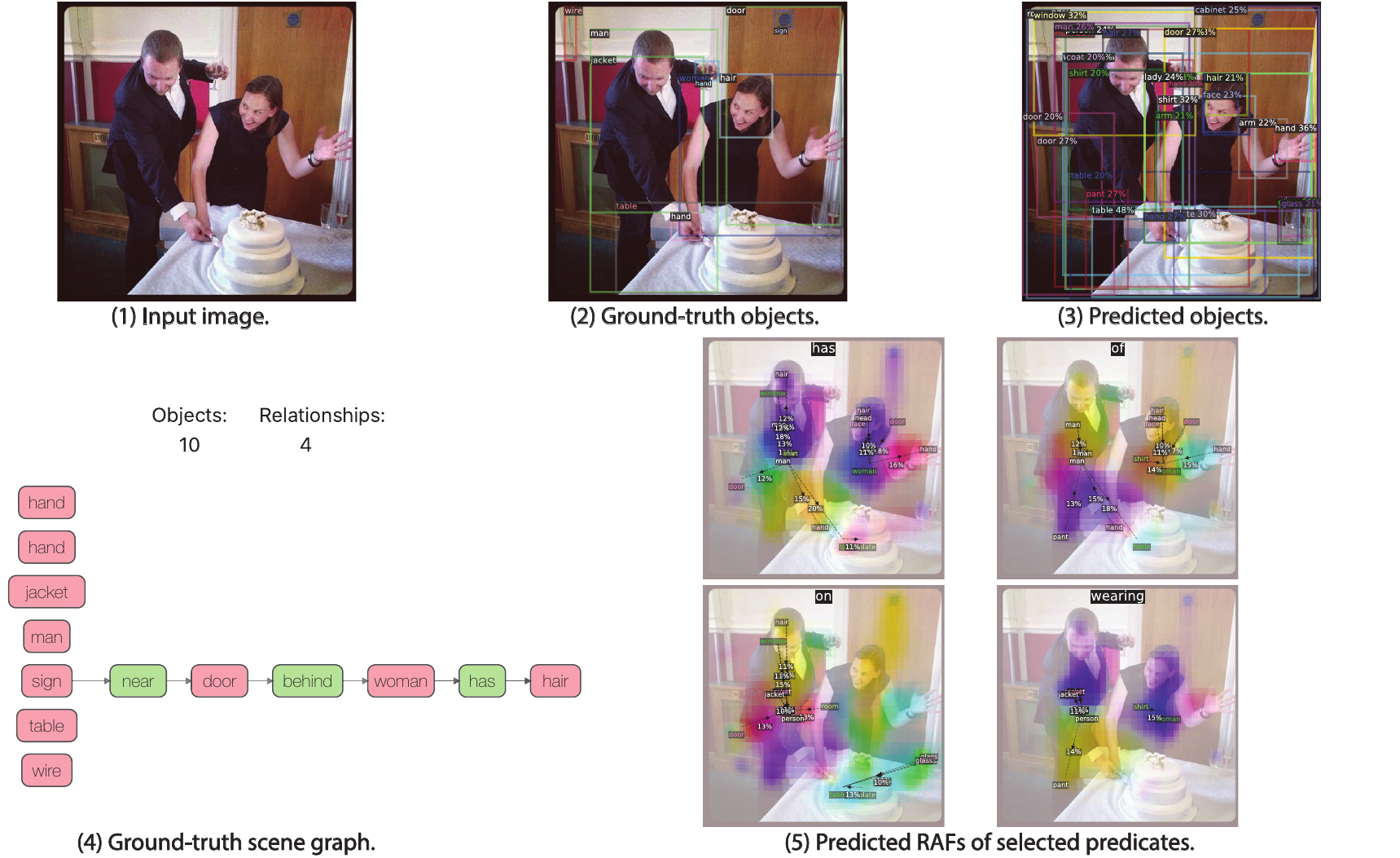}
         \caption{Visualizations of test image \texttt{2342618.jpg}.}
         \label{fig:vis_e}
     \end{subfigure}
     \hfill
     \begin{subfigure}[b]{\linewidth}
         \centering
         \includegraphics[width=\linewidth]{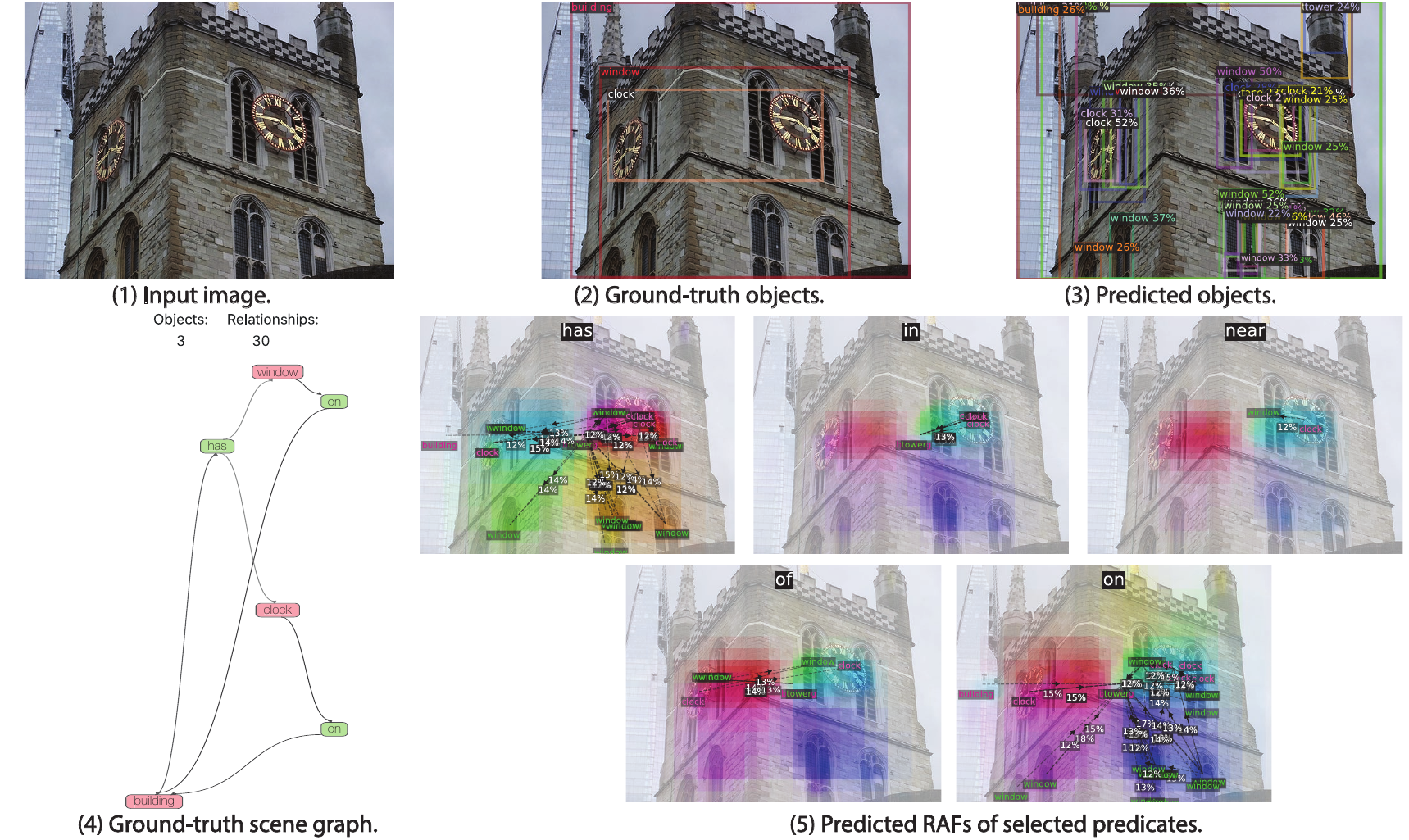}
         \caption{Visualizations of test image \texttt{2343159.jpg} (there are duplicate relationships in the annotation).}
         \label{fig:vis_f}
     \end{subfigure}
        \caption{Visualizations of results on VG-150 test set.}
        \label{fig:vis}
\end{figure*}

We then tested on wild images for examining the applicability of our approach. For testing predicate \texttt{WEARING}, we downloaded 3 online images containing \texttt{jacket} shown in Figure~\ref{fig:jacket} but in different scenarios: a product photo of \texttt{jacket} (\hyperref[fig:jacket]{4a}); \texttt{man} \texttt{WEARING} \texttt{jacket} (\hyperref[fig:jacket]{4b}); \texttt{jacket} \texttt{ON} \texttt{bed} (\hyperref[fig:jacket]{4c}). The RAF predictions for \texttt{WEARING} are visualized, and the maximum unnormalized RAF vector norm is $\mathrm{0.79}$, $\mathrm{4.02}$, and $\mathrm{0.67}$ for \hyperref[fig:jacket]{4a}, \hyperref[fig:jacket]{4b}, and \hyperref[fig:jacket]{4c}, respectively. The predicate \texttt{WEARING} gets the strongest response in Figure~\hyperref[fig:jacket]{4b} among the three images, demonstrating that RAFs successfully capture the semantics. It also exposes the problem of learning bias, such that even no person presents in \hyperref[fig:jacket]{4a} and \hyperref[fig:jacket]{4c}, there are responses of \texttt{WEARING} since \texttt{jacket} and \texttt{WEARING} often coexist.

\end{appendices}

\end{document}